%% file: main.tex
\documentclass[10pt]{article} 
\usepackage[preprint]{tmlr}
\input{math_commands.tex}

\usepackage{hyperref}
\usepackage{url}
\usepackage{tikz}
\usepackage{xcolor}
\usepackage[linesnumbered,ruled,vlined]{algorithm2e}

\usepackage{booktabs}      
\usepackage{amsmath,amssymb} 
\usepackage{amsfonts}       
\usepackage{nicefrac}       
\usepackage{microtype}      
\usepackage{xspace}
\usepackage{mathabx}
\usepackage{dsfont}
\usepackage{comment}
\usepackage{multirow}
\usepackage{adjustbox}
\usepackage{float}
\usepackage{wrapfig}
\usepackage{array}
\usepackage[dvipsnames]{xcolor}
\usepackage[table]{xcolor}
\usepackage{graphicx}
\usepackage{subcaption}
\usetikzlibrary{shapes.geometric}
\usetikzlibrary{tikzmark}

\makeatletter
\DeclareRobustCommand\onedot{\futurelet\@let@token\@onedot}
\def\@onedot{\ifx\@let@token.\else.\null\fi\xspace}
\def\eg{\emph{e.g}\onedot} 
\def\ie{\emph{i.e}\onedot}

\newcommand*\bigcdot{\mathpalette\bigcdot@{.5}}
\newcommand*\bigcdot@[2]{\mathbin{\vcenter{\hbox{\scalebox{#2}{$\m@th#1\bullet$}}}}}
\makeatother

\newcommand*\circled[1]{\tikz[baseline=(char.base)]{
            \node[shape=circle,draw,inner sep=1pt] (char) {\tiny {#1}};}}
\newcommand*\squared[1]{\tikz[baseline=(char.base)]{
            \node[shape=diamond,draw,inner sep=0.5pt] (char) {\tiny {#1}};}}

\newcommand*\tightsquare[1]{\tikz[baseline=(char.base)]{%
    \node[shape=regular polygon,regular polygon sides=4,draw,inner sep=0.3pt] (char) {\tiny {#1}};}}

\title{Modular Diffusion Models for Structured Visual Recognition}

\author{\name Siddhesh Khandelwal
\email skhandel@cs.ubc.ca \\
\addr Department of Compute Science, University of British Columbia\\
\addr Vector Institute for AI
\AND
\name Björn Ommer
\email b.ommer@lmu.de \\
\addr CompVis, Ludwig Maximilian University of Munich \\
\addr Munich Center for Machine Learning
\AND
\name Leonid Sigal
\email lsigal@cs.ubc.ca \\
\addr Department of Computer Science, University of British Columbia \\
\addr Vector Institute for AI \\
\addr CIFAR AI Chair
}

\def\month{MM}  
\def\year{YYYY} 
\def\openreview{\url{https://openreview.net/forum?id=XXXX}} 

\begin{document}

\maketitle

\begin{abstract} 
Traditional supervised methods for structured visual recognition tasks -- such as object detection, segmentation, and scene graph generation -- often produce deterministic, fixed outputs, limiting their ability to capture the inherent uncertainty in complex visual scenes. As a consequence, such point estimates are unable to capture the prediction uncertainty (or multi modality) intrinsic to these problems, often arising from natural ambiguities (\eg, ambiguity in size of partially occluded objects, local ambiguity of exact segmentation boundary, etc.) as well as noise and sparsity of training data. To address this limitation, we present Modular Diffusion Models (MDMs), a simple and novel framework that learns a distribution over structured outputs for a given input image. MDMs decompose the diffusion process into distinct, task-specific modules, each focused on capturing a different aspect of the structured information space, such as object categories, spatial locations, and inter-object relationships. This modular design allows each component to be learned independently, with seamless integration at inference without additional training. Furthermore, the modularity of MDMs enables the diffusion process to easily operate over the heterogeneous output space common in many structured learning tasks (\eg, a continuous bounding boxes and discrete class labels). Experimental results over three distinct structured tasks -- object detection, instance segmentation, and scene graph generation -- highlight the benefits of our proposed framework.
\end{abstract}

\input{sections/0_introduction}
\input{sections/1_relatedwork}

\input{sections/2_background}
\input{sections/3_approach}
\input{sections/4_experiments}

\section{Conclusion}
In this work, we propose Modular Diffusion Models (MDMs), a general framework for learning complex distributions in structured learning tasks. Instead of modeling the joint distribution over heterogeneous components directly, our approach factorizes it into a sequence of simpler conditional distributions. We provide a simple transformer-based instantiation of the proposed framework, and highlight the generality and effectiveness of MDMs through extensive experiments across three distinct structured learning problems.

\textbf{Ethical Considerations. }The proposed MDM framework is a general-purpose approach for modeling structured distributions and is not tied to any particular application domain. We argue that improved uncertainty modeling can be beneficial in many vision tasks by providing a richer representation of model predictions and ambiguity. However, as is the case with any AI approach, the framework could be applied in settings such as surveillance, monitoring, or automated decision-making, where prediction errors or biased model behavior may have adverse consequences. Additionally, machine learning models trained on real-world data can inherit and amplify biases present in the training datasets. For example, under representation of certain object categories, environments, or demographics may lead to uneven performance across different groups. This behavior is not unique to MDM and is a challenge faced by many AI systems. Finally, our work does not involve the collection of new human-subject data, and all experiments are conducted using publicly available benchmark datasets commonly used in computer vision research \citep{lin2014microsoft, krishna2017visual}.

\section{Acknowledgments and Disclosure of Funding}
This work was funded, in part, by the Vector Institute for AI, Canada CIFAR AI Chair, NSERC Canada Research Chair (CRC), and NSERC Discovery Grant. Resources used in preparing this research were provided, in part, by the Province of Ontario, the Government of Canada through CIFAR, the Digital Research Alliance of Canada, companies\footnote{https://vectorinstitute.ai/\#partners} sponsoring the Vector Institute, and Advanced Research Computing at the University of British Columbia. Additional hardware support was provided by John R. Evans Leaders Fund CFI grant and Digital Research Alliance of Canada under the Resource Allocation Competition award.

\begin{figure}[t]
\centering
\begin{subfigure}{\textwidth}
\centering

\begin{minipage}{0.32\textwidth}
    \centering
    \includegraphics[width=\textwidth]{images/inst_examples/original_000000149622_0.jpg}
\end{minipage}
\hfill
\begin{minipage}{0.64\textwidth}
    \centering

    \begin{minipage}{0.49\textwidth}
        \includegraphics[width=\textwidth]{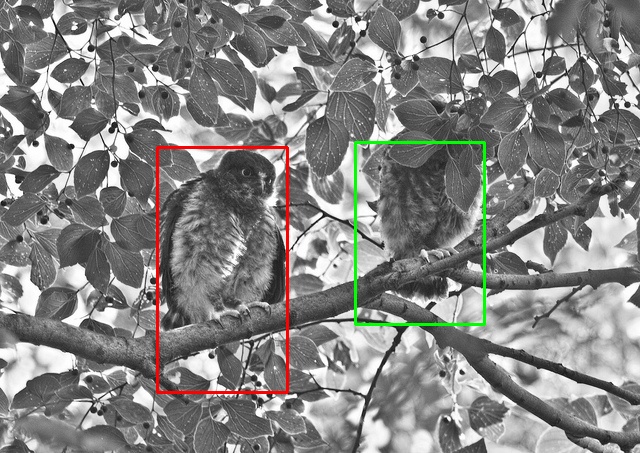}
    \end{minipage}
    \hfill
    \begin{minipage}{0.49\textwidth}
        \includegraphics[width=\textwidth]{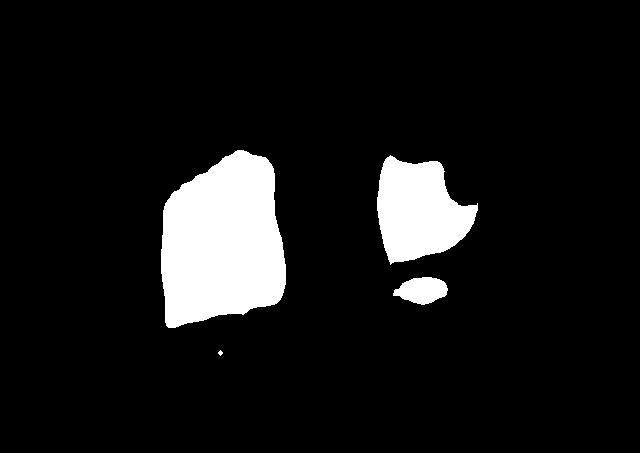}
    \end{minipage}

    \vspace{0.5em}

    \begin{minipage}{0.49\textwidth}
        \includegraphics[width=\textwidth]{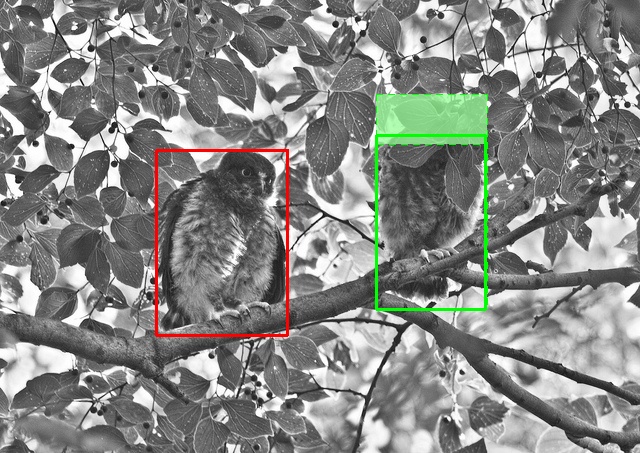}
    \end{minipage}
    \hfill
    \begin{minipage}{0.49\textwidth}
        \includegraphics[width=\textwidth]{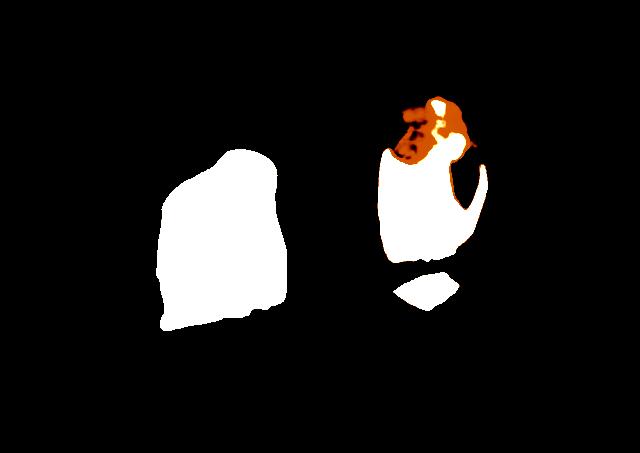}
    \end{minipage}

\end{minipage}

\caption{The ``bird'' object on the right is partially occluded by tree leaves. Unlike the deterministic Mask-RCNN \citep{he2017mask} (top), MDM (bottom) captures ambiguity in both the ``bird’s'' head location (\textcolor{green}{green}) and its shape (darker regions in the heatmap).}
\label{fig:inst-seg-a}
\end{subfigure}

\vspace{1em}

\begin{subfigure}{\textwidth}
\centering

\begin{minipage}{0.32\textwidth}
    \centering
    \includegraphics[width=\textwidth]{images/inst_examples/original_000000062353_0.jpg}
\end{minipage}
\hfill
\begin{minipage}{0.64\textwidth}
    \centering

    \begin{minipage}{0.49\textwidth}
        \includegraphics[width=\textwidth]{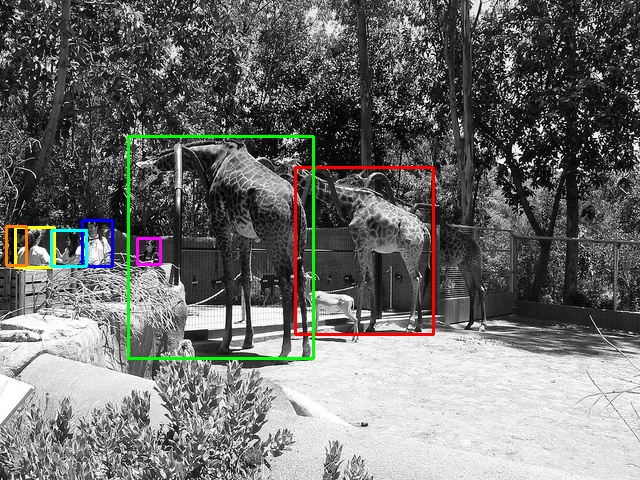}
    \end{minipage}
    \hfill
    \begin{minipage}{0.49\textwidth}
        \includegraphics[width=\textwidth]{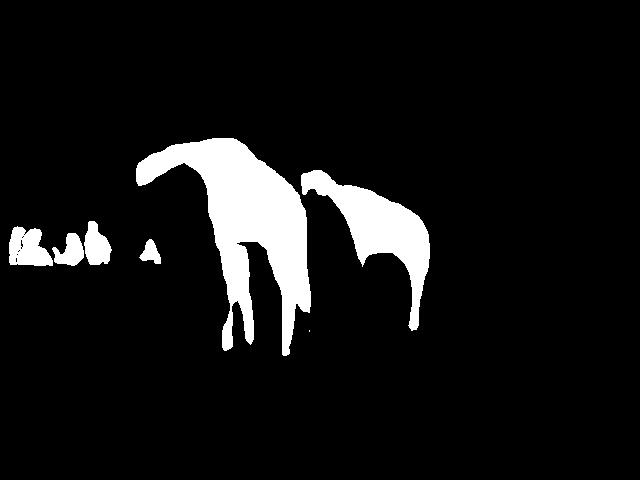}
    \end{minipage}

    \vspace{0.5em}

    \begin{minipage}{0.49\textwidth}
        \includegraphics[width=\textwidth]{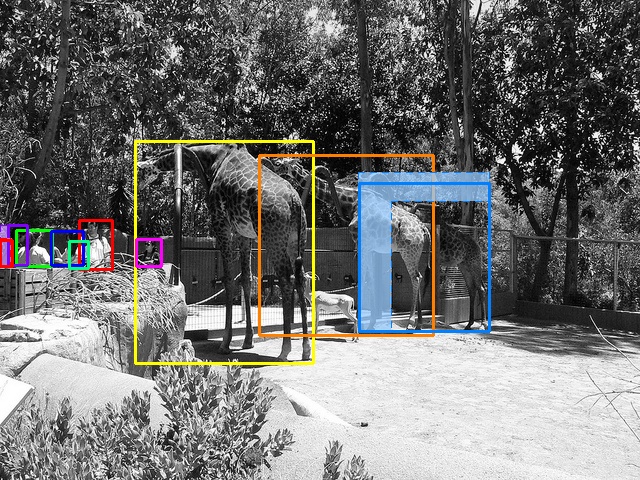}
    \end{minipage}
    \hfill
    \begin{minipage}{0.49\textwidth}
        \includegraphics[width=\textwidth]{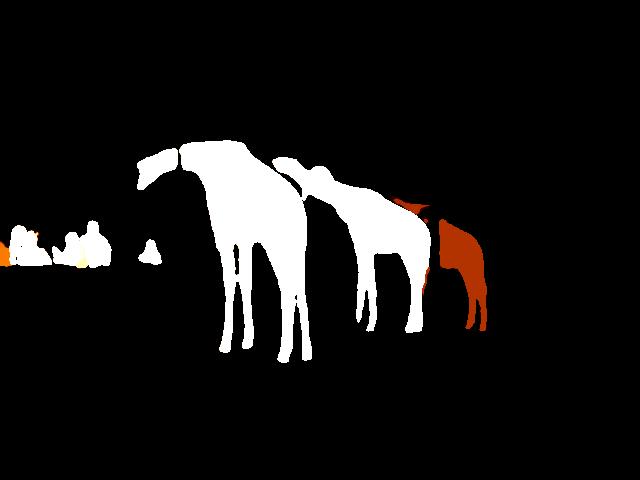}
    \end{minipage}

\end{minipage}

\caption{The rightmost ``giraffe'' is partially occluded by others. Unlike the deterministic Mask R-CNN \citep{he2017mask} (top), MDM (bottom) captures uncertainty in localization due to the ``giraffe’s'' upper body (\textcolor{NavyBlue}{blue}).}
\label{fig:inst-seg-b}
\end{subfigure}

\caption*{}
\end{figure}

\begin{figure}[btp]
\ContinuedFloat
\centering

\begin{subfigure}{\textwidth}
\centering

\begin{minipage}{0.32\textwidth}
    \centering
    \includegraphics[width=\textwidth]{images/inst_examples/original_000000080153_0.jpg}
\end{minipage}
\hfill
\begin{minipage}{0.64\textwidth}
    \centering

    \begin{minipage}{0.49\textwidth}
        \includegraphics[width=\textwidth]{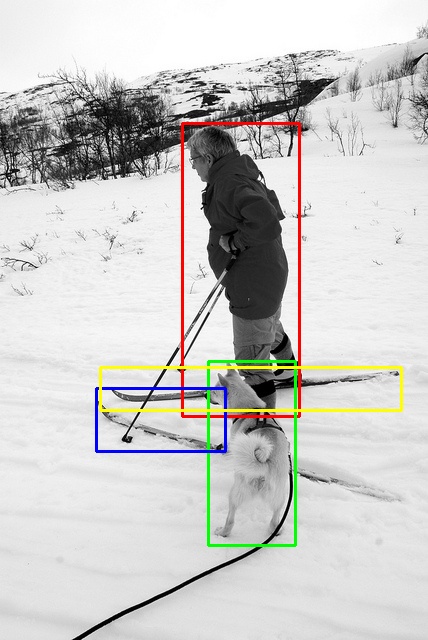}
    \end{minipage}
    \hfill
    \begin{minipage}{0.49\textwidth}
        \includegraphics[width=\textwidth]{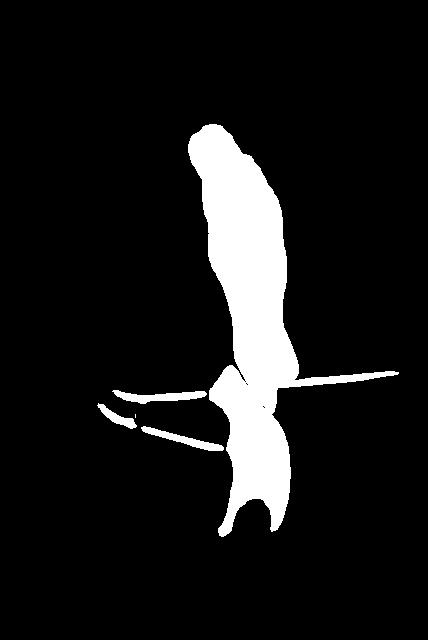}
    \end{minipage}

    \vspace{0.5em}

    \begin{minipage}{0.49\textwidth}
        \includegraphics[width=\textwidth]{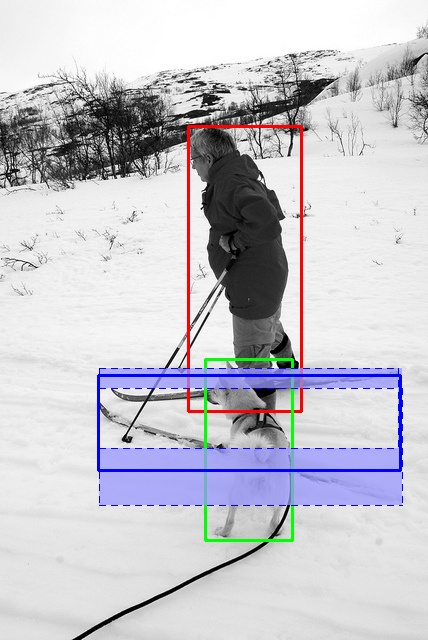}
    \end{minipage}
    \hfill
    \begin{minipage}{0.49\textwidth}
        \includegraphics[width=\textwidth]{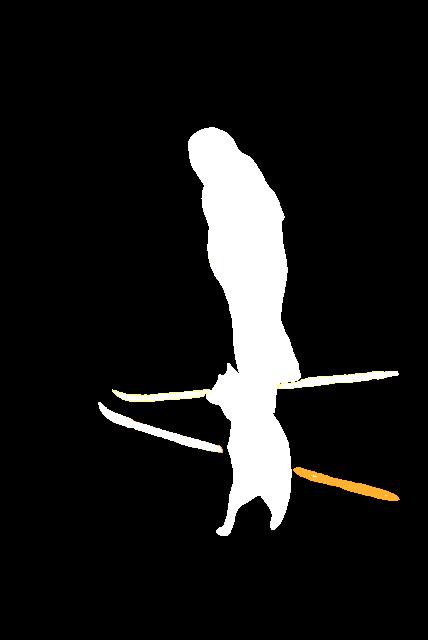}
    \end{minipage}

\end{minipage}
\caption{The tail of the left ``ski'' is covered by the snow. Unlike the deterministic Mask R-CNN \citep{he2017mask} (top), MDM (bottom) captures uncertainty in both localization (\textcolor{blue}{blue}) and shape (darker regions in the heatmap).}

\label{fig:inst-seg-c}
\end{subfigure}
\caption{\textbf{Uncertainty analysis.} Results shown on the task of instance segmentation. We contrast predictions from a deterministic Mask R-CNN \citep{he2017mask} model with those from our proposed MDM framework. For each image, MDM generates 1000 samples and aggregates masks into a heatmap; brighter regions indicate consistent predictions across samples, while darker regions reflect uncertainty. Variability in object localization is also highlighted by plotting the mean bounding box and shading the region spanning the minimum and maximum predicted box extents; larger shaded regions indicate greater uncertainty.}
\label{fig:inst-seg}
\end{figure}

\begin{figure}[btp]
\centering
    \begin{subfigure}{0.95\textwidth}
        \centering
        \includegraphics[width=\textwidth]{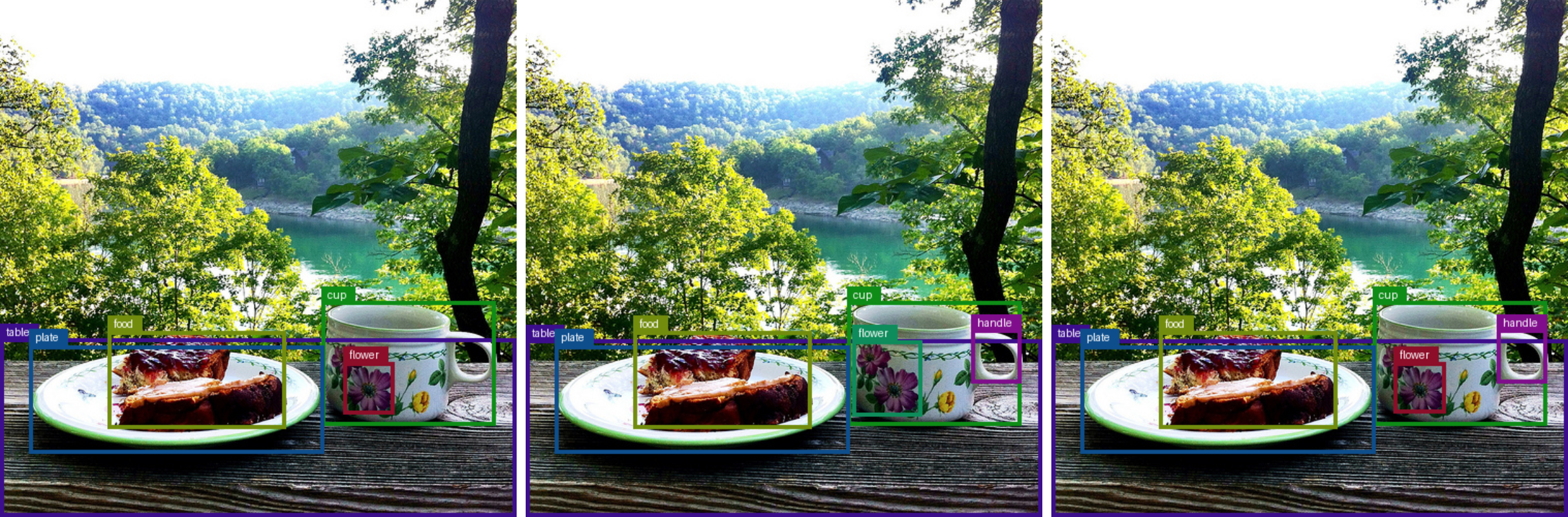}
        \centering
        \includegraphics[width=0.6\textwidth]{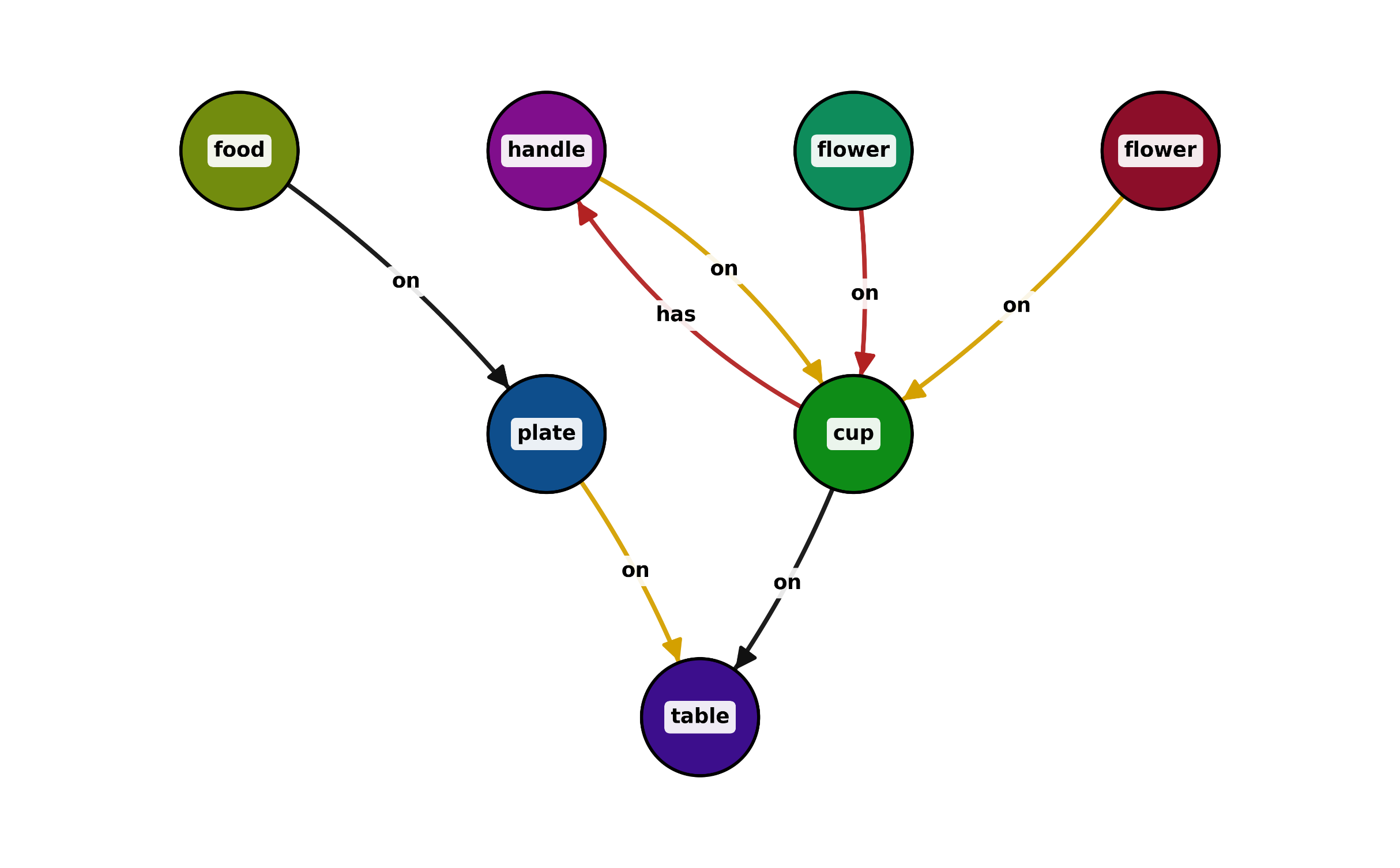}
        \caption{}
        \vspace{1cm}
    \end{subfigure}
    
    \begin{subfigure}{0.95\textwidth}
        \centering
        \includegraphics[width=\textwidth]{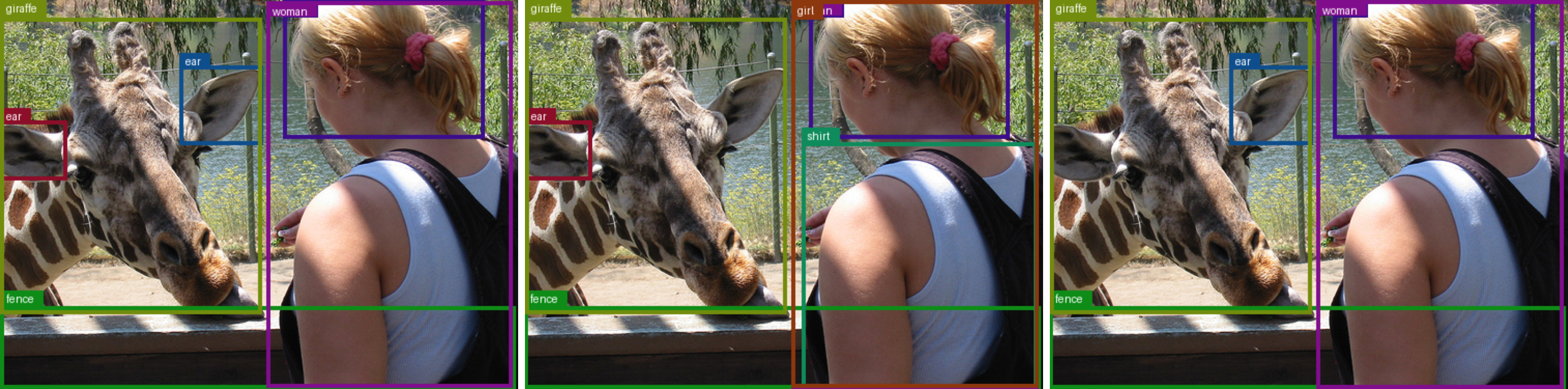}
        \centering
        \includegraphics[width=\textwidth]{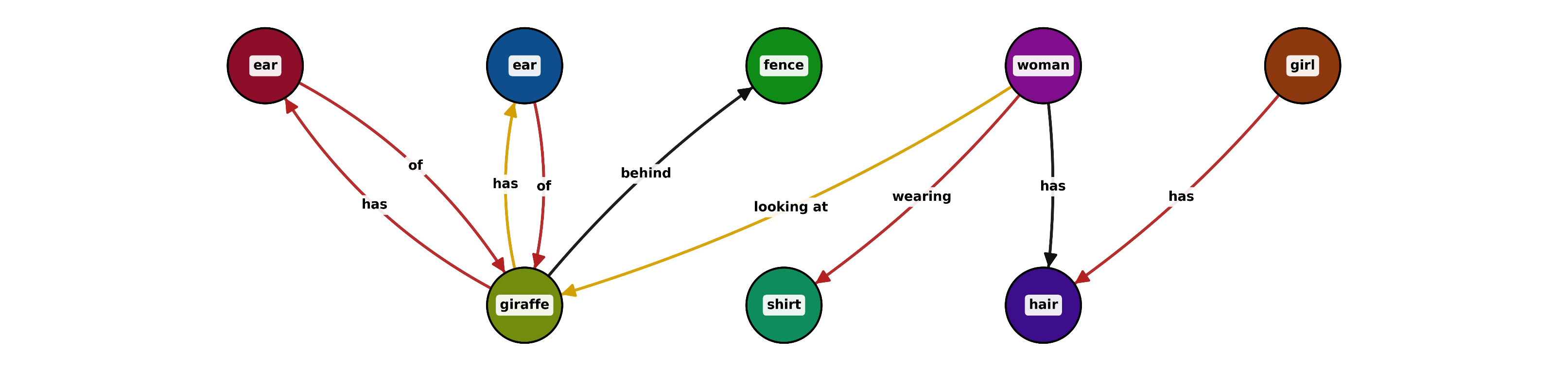}
        \caption{}
    \end{subfigure}
    
    \caption*{}
\end{figure}

\begin{figure}[btp]
\ContinuedFloat
\centering
\begin{subfigure}{0.95\textwidth}
        \centering
        \includegraphics[width=\textwidth]{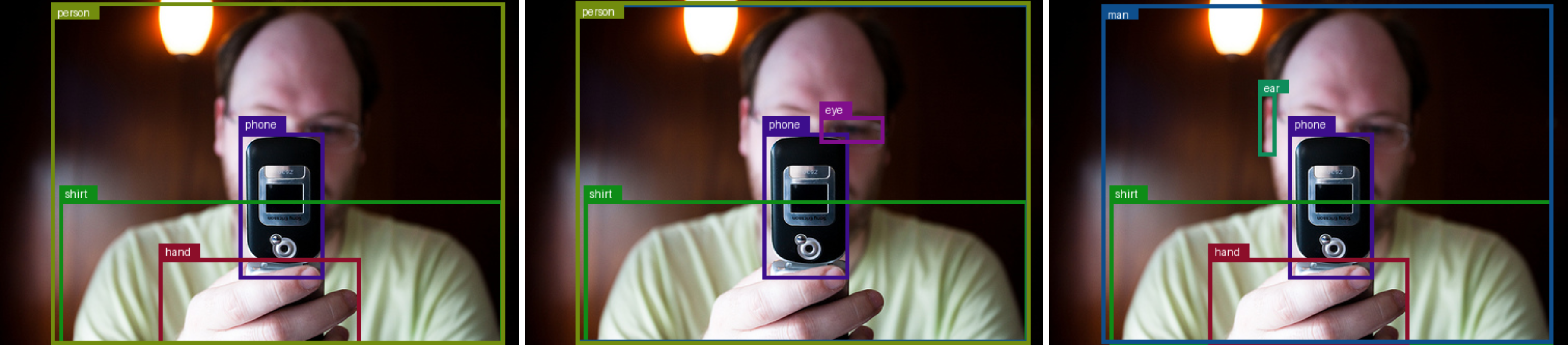}
        \centering
        \includegraphics[width=\textwidth]{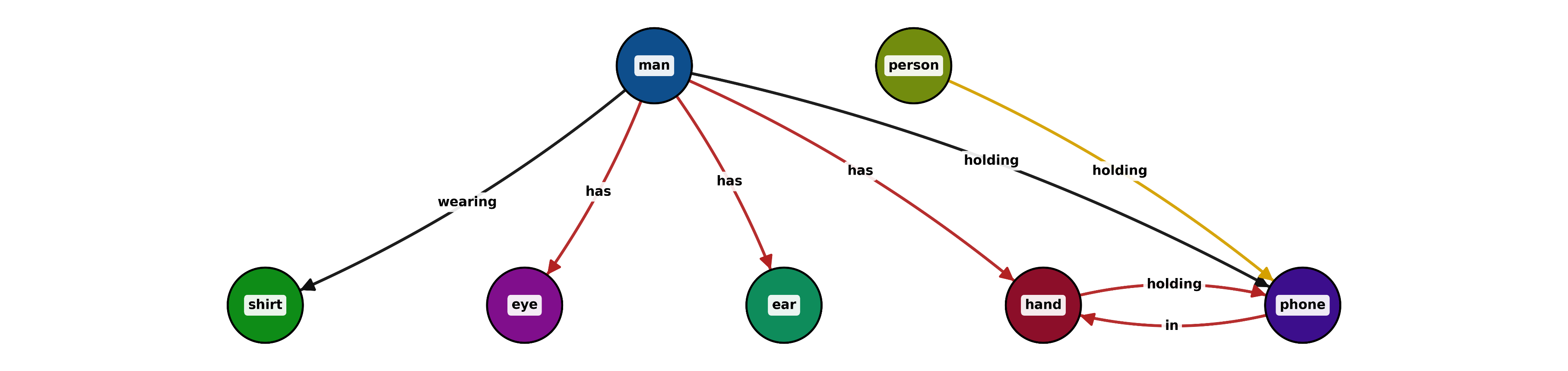}
        \caption{}
\end{subfigure}
\caption{\textbf{Qualitative Analysis.} Results for the scene graph generation task. We sample three outputs from our MDM model and visualize the aggregated scene graph together with the corresponding bounding boxes. \textbf{Black} edges denote triplets that appear in all three samples, \textcolor{YellowOrange}{\textbf{yellow}} edges denote triplets that appear in exactly two samples, and \textcolor{red}{\textbf{red}} edges denote triplets that are unique to a single sample. MDM is able to generate distinct but plausible graphs given the same input image.}
\label{fig:sg}
\end{figure}

\clearpage
\bibliography{main}
\bibliographystyle{tmlr}

\clearpage
\appendix
\section{Appendix}
\subsection{Transformer Architecture for Scene Graph Generation}
\label{appendix:sg-arch}
In Section \ref{sec:exp-sg} of the main paper we briefly describe that for the task of scene graph generation, we instantiate the MDM framework with a denoising decoder consisting of three components -- a subject decoder, an object decoder, and a predicate decoder. In this section we detail how the transformer-based formulation described in Section \ref{sec:tranformer-denoising} is extended to this three-decoder setup without introducing any additional task-specific architectural components. 

As described in Section \ref{sec:mdm-sg}, our proposed MDM framework for scene graph generation factorizes the joint distribution into two conditional distributions -- $\Pr(\{(\mathbf{b}_i^s, \mathbf{b}_i^o)\} \, |\, \mathbf{I})$ (proposal diffusion) and $\Pr(\{(\mathbf{l}_i^s, \mathbf{l}_i^o, \mathbf{l}_i^p)\} \, |\, \{(\mathbf{b}_i^s, \mathbf{b}_i^o)\}, \mathbf{I})$ (triplet diffusion) (Eq. (\ref{eq:scenegraphfactorization})). The forward diffusion processes for these conditionals independently corrupt the corresponding components as described in Section \ref{sec:mdm-sg}.

Naively modeling the denoising process for the aforementioned conditionals makes learning challenging due to the combinatorial nature of the output space. For example, in triplet diffusion, the output space has a cardinality of $\mathcal{N}_o \times \mathcal{N}_o \times \mathcal{N}_p$, where $\mathcal{N}_o$ and $\mathcal{N}_p$ are the number of entity and predicate categories respectively. 

Therefore, to make learning tractable while still allowing the subject, object, and predicate components to depend on each other, we follow an approach similar to \cite{khandelwal2022iterative}, and further factorize each denoising process. Concretely,
\begin{align}
    \Pr\Bigl(\{(\mathbf{b}_i^s, \mathbf{b}_i^o)\} \,\Big|\, \mathbf{I}\Bigr)
    = \;
    \underbrace{
    \Pr\Bigl(\{\mathbf{b}_i^s\} \,\Big|\, \mathbf{I}\Bigr)
    }_{\text{subject decoder}}
    \cdot
    \underbrace{
    \Pr\Bigl(\{\mathbf{b}_i^o\} \,\Big|\, \{\mathbf{b}_i^s\}, \mathbf{I}\Bigr)
    }_{\text{object decoder}}
\end{align}

\begin{align}
    \begin{split}
        \Pr\Bigl(\{(\mathbf{l}_i^s, \mathbf{l}_i^o, \mathbf{l}_i^p)\} \,\Big|\, \{(\mathbf{b}_i^s, \mathbf{b}_i^o)\}, \mathbf{I}\Bigr)
        = \;&
        \underbrace{
        \Pr\Bigl(\{\mathbf{l}_i^s\} \,\Big|\, \{(\mathbf{b}_i^s, \mathbf{b}_i^o)\}, \mathbf{I}\Bigr)
        }_{\text{subject decoder}}
        \\
        \cdot&
        \underbrace{
        \Pr\Bigl(\{\mathbf{l}_i^o\} \,\Big|\, \{\mathbf{l}_i^s\}, \{(\mathbf{b}_i^s, \mathbf{b}_i^o)\}, \mathbf{I}\Bigr)
        }_{\text{object decoder}}
        \\
        \cdot&
        \underbrace{
        \Pr\Bigl(\{\mathbf{l}_i^p\} \,\Big|\, \{(\mathbf{l}_i^s, \mathbf{l}_i^o)\}, \{(\mathbf{b}_i^s, \mathbf{b}_i^o)\}, \mathbf{I}\Bigr)
        }_{\text{predicate decoder}}
    \end{split}
\end{align}

Each of these three decoders ($r \in \{s, o, p\}$) is architecturally identical to the formulation introduced in Section \ref{sec:tranformer-denoising} and processes its respective noisy input (subject/object/predicate box or label). Note that in proposal diffusion, given noisy subject and object boxes $\{({\mathbf{b}}_{i, t}^s, {\mathbf{b}}_{i,t}^o)\}$, the noisy predicate box $\mathbf{b}_{i, t}^p$ is defined as the bounding box formed by the centers of $\mathbf{b}_{i, t}^s$ and $\mathbf{b}_{i, t}^o$. The input queries $\{\mathbf{q}_{i}^r\}; r \in \{s, o, p\}$ are subsequently obtained by following the formulation in Eq. (\ref{eq:input-projection}). 

For a particular layer $m$, the query features $\{\mathbf{q}_{i}^{s,k,m}\}$ for the subject decoder $s$ are computed by applying the formulation in Eq. (\ref{eq:time-projection}). For the object decoder $o$, we modulate the object queries based on the subject queries as follows,
\begin{align}
    \begin{split}
        \boldsymbol{\rho}_i^{o,m}, \boldsymbol{\tau}_i^{o,m} &= f_{o,m}(\mathbf{q}_{i}^{s,k,m})\\
        \mathbf{q}_{i}^{o,k,m-1} &\leftarrow \mathbf{q}_{i}^{o,k,m-1} \odot (1 + \boldsymbol{\rho}_i^{o,m}) + \boldsymbol{\tau}_i^{o,m}
    \end{split}
\end{align}
The updated features $\{\mathbf{q}_{i}^{o,k,m}\}$ are then obtained by applying the formulation in Eq. (\ref{eq:time-projection}). Finally, the predicate queries are modulated using both subject and object queries,
\begin{align}
    \begin{split}
        \boldsymbol{\rho}_i^{p,m}, \boldsymbol{\tau}_i^{p,m} &= f_{p,m}(\mathbf{q}_{i}^{s,k,m}, \mathbf{q}_{i}^{o,k,m})\\
        \mathbf{q}_{i}^{p,k,m-1} &\leftarrow \mathbf{q}_{i}^{p,k,m-1} \odot (1 + \boldsymbol{\rho}_i^{p,m}) + \boldsymbol{\tau}_i^{p,m}
    \end{split}
\end{align}
The updated query features $\{\mathbf{q}_{i}^{p,k,m}\}$ are then similarly obtained by following Eq. (\ref{eq:time-projection}). Crucially, it should be noted that there are no structural or architectural differences across the three decoders. The inherent dependencies between the subject, object, and predicate components are established entirely via the query modulation steps described above.

\end{document}

%% file: math_commands.tex
\usepackage{amsmath,amsfonts,bm}

\def\eqref#1{equation~\ref{#1}}

\def\1{\bm{1}}

\DeclareMathAlphabet{\mathsfit}{\encodingdefault}{\sfdefault}{m}{sl}
\SetMathAlphabet{\mathsfit}{bold}{\encodingdefault}{\sfdefault}{bx}{n}



%% file: sections/0_introduction.tex
\section{Introduction}
\label{sec:intro}

\begin{figure}[t]
    \centering
    \includegraphics[width=0.95\textwidth]{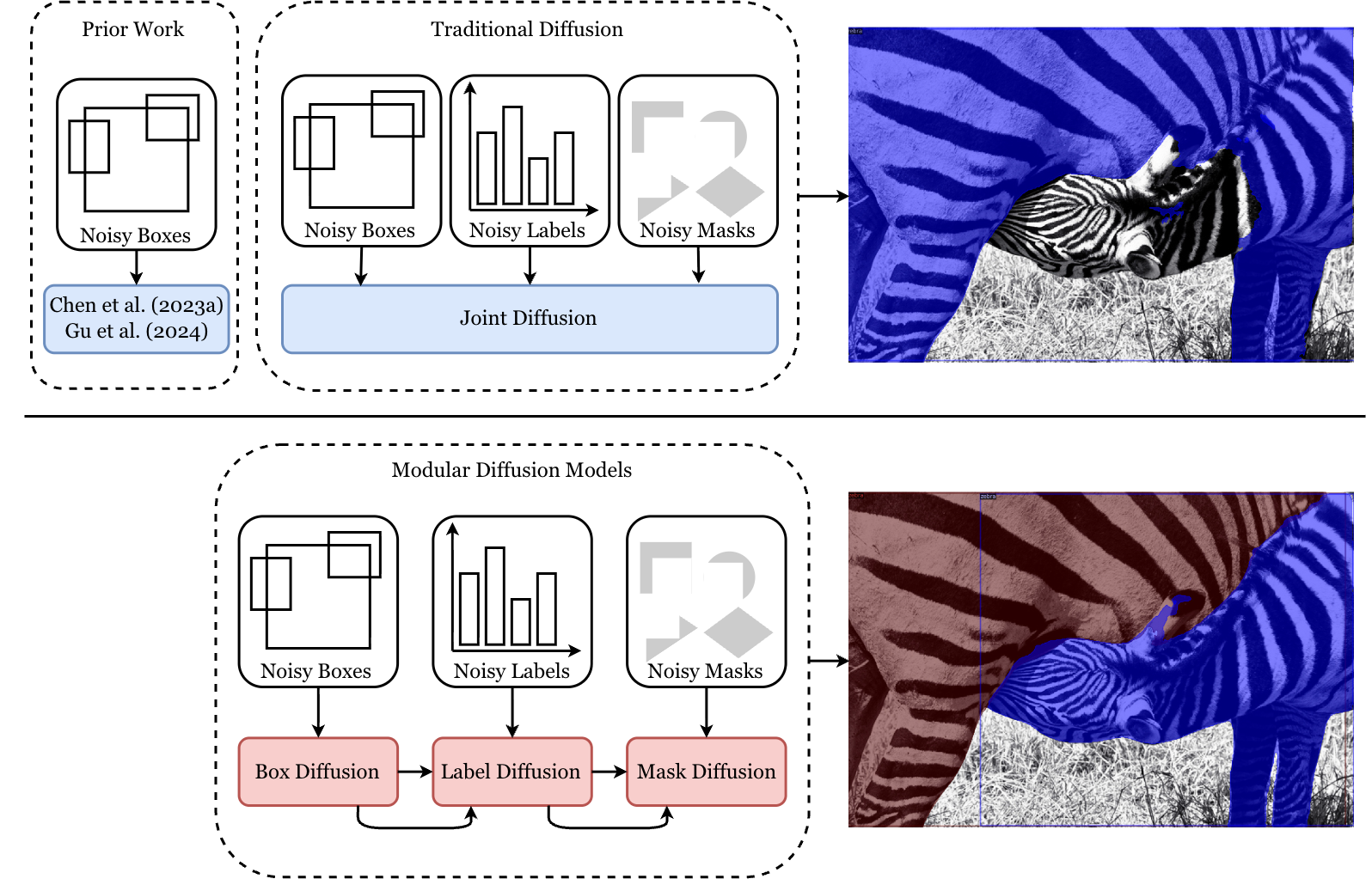}
    \caption{\textbf{Modular Diffusion Models} factorize the complex joint structured distribution into several easy-to-estimate conditionals, each modeled using a diffusion process. In contrast to existing work (top), MDM (bottom) models a diffusion process over all components of the structured distribution. Additionally, compared to a naive joint approach (top), MDM preserves cross-component alignment during denoising, avoiding incoherent intermediate states and enabling more reliable reconstruction of structured outputs. In the example shown above, MDM is able to correctly identify and accurately segment the two zebras in the scene.}
    \label{fig:mdm-teaser}
\end{figure}

Structured visual recognition tasks -- such as object detection \citep{ren2015faster,girshick2015fast,uijlings2013selective,redmon2016you,carion2020end}, segmentation \citep{he2017mask,carion2020end,cheng2022masked,li2023mask}, and scene graph generation \citep{desai2021learning, newell2017pixels, qi2019attentive, tang2020unbiased, tang2019learning, xu2017scene, yang2018graph, zareian2020bridging, zellers2018neural} -- require models to capture different levels of detail and context within a visual scene ({\em e.g.}, an image or video) using structured representations. These representations range from simple bounding boxes that localize objects in object detection, to graphical structures in scene graph generation, where objects and their corresponding interactions are represented as nodes and edges respectively. These structures are fundamental to methods in areas like medical imaging, autonomous driving, and security systems. They are also frequently used as intermediate representation in a variety of complex downstream problems, and have proven effective in improving performance on tasks such as image captioning \citep{gu2019unpaired, yang2019auto, nguyen2021defense}, visual question answering \citep{hudson2019gqa, tang2019learning, lee2019visual}, and image generation \citep{johnson2018image, li2019pastegan}.

Learning these representations can be interpreted as modeling the conditional distribution over the desired type of structured information (\eg segmentation masks, or scene graphs) given an input image (or video). Classical approaches to structured visual understanding were often explicitly probabilistic, utilizing graphical models such as Markov random fields and conditional random fields \citep{quattoni2004conditional, held1997markov, lafferty2001conditional, boykov2006graph}. These methods explicitly represented uncertainty and dependencies between output variables, allowing multiple plausible interpretations of a scene to be expressed through a structured probability distribution. However, estimating such distributions is inherently challenging due to the combinatorial growth of the output space. For instance, in scene graph generation, the number of possible graphs grows combinatorially with the number of objects and interactions involved.

To avoid the aforementioned issue, recent deep learning-based approaches often approximate the conditional distribution by learning to predict the distribution mean (a point estimate) \citep{dong2021visual, liu2021fully, newell2017pixels, tang2020unbiased, xu2017scene, yang2018graph, zareian2020bridging, suhail2021energy, khandelwal2022iterative}, usually by employing supervised learning techniques on an annotated training set. While this shift has led to substantial gains in accuracy and scalability, it has also resulted in models that produce a single, fixed structured output for each visual input. As a consequence, modern approaches often struggle to represent inherent ambiguities present in complex visual scenes.

Learning generative models to approximate complex distributions is a well-established problem in the domain of image generation \citep{kingma2018glow, razavi2019generating, brock2018large, karras2017progressive, goodfellow2014generative}. Diffusion-based processes \citep{ho2020denoising, rombach2022high, ruiz2023dreambooth, nair2023steered, saharia2022photorealistic}, in particular, have shown impressive ability to generate realistic, high-resolution images, both unconditionally and under the guidance of textual prompts. Diffusion processes fundamentally consists of two stages -- (i) a forward phase, where noise is incrementally added to the input data, and (ii) a denoising phase, during which a neural network learns to gradually denoise a noisy input. Extending this framework to structured recognition tasks, however, is uniquely challenging due to the heterogeneous and interdependent nature of the output space. 

For instance, in the case of instance segmentation, the desired structured result includes continuous outputs (bounding boxes for individual objects ; \eg ``person'' and ``bicycle'') and discrete outputs (corresponding object labels and spatial segmentation masks). These components are inherently coupled -- the object category often determines the expected size and shape of its bounding box and mask, while, in turn, the geometry of the mask and the spatial placement of the box provide important cues for identifying the object label.

A naive extension of diffusion models to structured prediction would involve defining a joint diffusion process over all output components, independently corrupting each element and training a model to jointly denoise them. However, modeling such complex structured distributions is challenging, particularly when they combine discrete (\eg class labels) and continuous elements (\eg bounding boxes and segmentation masks). Using the aforementioned example of instance segmentation, independently noising object bounding boxes, class labels, and segmentation masks can destroy the alignment between these components, such that they no longer correspond to the same underlying entities. For example, a segmentation mask originally associated with a ``person'' may become paired with a label such as ``bicycle'', or with a mismatched bounding box that no longer tightly encloses it. These inconsistencies produce incoherent intermediate states, making denoising substantially more difficult. This is shown qualitatively in Figure \ref{fig:mdm-teaser} and quantitatively in Table \ref{tab:coco-inst-ablation}. Modeling the aforementioned joint diffusion process leads to a loss in performance across multiple denoising steps, suggesting the inability of the model to accurately capture the underlying structured data distribution.

As a result, existing diffusion-based approaches typically avoid full joint modeling and instead operate on simplified or homogeneous representations. For instance, in the diffusion-based object detection model proposed in \citet{chen2023diffusiondet}, the forward phase only applies noise to the bounding box coordinates, while generating class labels is treated as an auxiliary task, which is learned through an additional head in the denoising model. Consequently, such an approach limits the model's capacity to learn a joint distribution over all components, resulting in fixed estimations for parts of the structured representation.

In this work, we introduce Modular Diffusion Models (MDMs), a novel framework designed to effectively learn distributions over structured outputs. The central concept of our approach is to decompose the joint conditional distribution, parameterized by the diffusion process, 
into a series of conditional distributions -- each over a homologous portion of the output space. Each of these conditional distributions, we parameterize using a separate diffusion process, resulting in the complete model being a modular chain of diffusion processes. 
For instance, in the task of instance segmentation, rather than attempting to learn a single diffusion process that concurrently denoises bounding boxes, class labels, and segmentation masks, our proposed MDM framework independently learns separate diffusion processes for each of these components. The benefits of such a modular factorization are manyfold - (i) it simplifies the learning process by allowing individual modules to concentrate solely on denoising a single homogeneous input; (ii) it effectively accommodates heterogeneous inputs, as each diffusion module can be specifically designed to work with either continuous or categorical data; and (iii) the independently trained diffusion processes can be seamlessly integrated during inference, enabling the generation of complete structured representations without requiring any additional training.

We instantiate the individual diffusion processes in the MDM framework using a simple transformer-based architecture that performs iterative denoising of structured components conditioned on the input image and ancestral context. Concretely, we construct input queries directly from noisy variables and modulate them with time-aware conditioning, enabling dynamic adaptation across diffusion steps. These queries are processed by a multi-layer transformer decoder, which progressively refines them through attention-based interactions to produce clean structured predictions. We highlight the effectiveness of our proposed model and framework by achieving strong performance across diverse structured visual recognition tasks -- object detection, instance segmentation, and scene graph generation -- without requiring task-specific specialized design components. Specifically, on object detection and instance segmentation, MDM outperforms existing diffusion-based approaches by up to $8.4\text{AP}^{\text{mask}}$ and $3.9\text{AP}^{\text{box}}$, while achieving parity with other strong deterministic approaches. Furthermore, on the task of scene graph generation, MDM outperforms the most competitive baselines by $2.6 / 1.6$ on $\text{hR@50/100}$. Beyond absolute performance metrics, contrary to determinstic models that are constrained to point estimates, MDM is able to capture the uncertainity in the underlying data distribution and sample multiple plausible structured outputs for a single image.

\vspace{0.1in}
\noindent
{\bf Contributions.} Our contributions can be summarized as follows -- (i) We introduce Modular Diffusion Models (MDMs), a general framework for modeling complex structured distributions by factorizing them into a sequence of simpler conditional diffusion processes; (ii) We propose a simple and general transformer-based instantiation of MDMs that performs iterative, context-aware denoising; (iii) We demonstrate the effectiveness of the framework across multiple structured visual recognition tasks with heterogeneous outputs, without requiring task-specific architectural components; (iv) We highlight the flexibility of MDMs, enabling fine-grained control over individual diffusion processes during inference without additional training.

%% file: sections/1_relatedwork.tex
\section{Related Work}
\label{sec:relatedwork}

\subsection{Structured Learning}
\noindent\textbf{Object Detection.} Object detection is a fundamental task in computer vision that involves identifying and localizing objects in an image. The field has seen substantial progress due to advancements in convolutional neural networks (CNNs), with most modern methods divided into \emph{one-stage} and \emph{two-stage} categories. Two-stage detectors, such as Faster R-CNN \citep{ren2015faster} and Fast R-CNN \citep{girshick2015fast}, first generate candidate box proposals -- either using external algorithms \citep{uijlings2013selective} or through a learned  proposal network (RPN) \citep{ren2015faster}. The second stage shares convolutional features across all region proposals, and generates refinements over the proposal boxes and corresponding labels. One-stage detectors were introduced to address the slower runtimes of two-stage models. These methods aim to simplify the process by detecting objects in a single pass \citep{redmon2016you, lu2016visual}. More recently, transformers-based approaches like DETR \citep{carion2020end} have gained significant popularity. These approaches formulate object detection as a set prediction problem, wherein boxes are generated via repeated self- and cross-attention \citep{vaswani2017attention} over fixed set of learnable queries, trained via a matching algorithm. Subsequent improvements in transformer-based methods have focused on enhancing convergence speeds and detection quality. For example, \citet{zhu2020deformable, liu2022dab} use anchor-boxes to guide image conditioning, and \citet{zhang2022dino, li2022dn} leverage denoising queries during training to improve learning stability and accelerate convergence.

\noindent
\noindent\textbf{Instance Segmentation.} Instance segmentation extends object detection by not only localizing objects but also generating precise pixel-level masks for each object instance. Early methods, such as Mask-RCNN \citep{he2017mask}, introduce an additional mask head that operates over proposed box aligned features, allowing simultaneous prediction of object classes and segmentation masks. Transformer-based approaches have recently seen increased use. DETR \citep{carion2020end} combines image-based cross-attention \citep{vaswani2017attention} with a convolutional up-scaling network to generate instance masks. Subsequent methods \citep{cheng2022masked, li2023mask} leverage multi-scale features from CNN backbones to directly generate masks from transformer queries via a channel-wise inner product between the query vectors and feature map at each spatial location, eliminating the need for separate upsampling networks.

\vspace{0.5em}
\noindent\textbf{Scene Graph Generation.} Scene graph generation focuses on modeling the relationships between object entities within an image, requiring methods that can both accurately detect objects and reason about their inter-object semantics. This task has been extensively studied in computer vision \citep{desai2021learning, khandelwal2021segmentation, li2021sgtr, liu2021fully, newell2017pixels, qi2019attentive, tang2020unbiased, tang2019learning, xu2017scene, yang2018graph, zareian2020bridging, zellers2018neural, khandelwal2022iterative, chen2023more, desai2022single, kim2024groupwise, fu2025hybrid}. As object detection is a critical component, existing approaches in this domain can also be broadly classified into two categories: \emph{one-stage} or \emph{two-stage} approaches. The initial phase of two-stage methods involves pre-training a strong object detector (like Faster-RCNN \citep{ren2015faster}) to obtain candidate entity representations (boxes and corresponding features) from an image. The second stage then involves training a predicate generation network, conditioned on these fixed entity representations. Existing works have explored various architectures for the predicate generation network, like Bi-directional LSTMs \citep{zellers2018neural}, Tree-LSTMs \citep{tang2019learning}, Graph Neural Networks \citep{yang2018graph}, and message-passing algorithms \citep{qi2019attentive, xu2017scene}. One-stage approaches instead generate both the entities and predicates jointly via end-to-end trainable modules, like fully convolutional networks \citep{liu2021fully} and transformer-based architectures \citep{cong2022reltr, dong2021visual, li2021sgtr, shit2022relationformer, desai2022single, kim2024groupwise, fu2025hybrid}. 

All the aforementioned methods generate deterministic structured estimates for a given input image. In contrast, our proposed MDM framework is able to learn a distribution, enabling sampling of multiple plausible structured representations.

\subsection{Generative Modeling}
\noindent\textbf{Diffusion Models.} Diffusion-based generative models have gained substantial popularity in recent years, largely due to their impressive performance on image generation tasks \citep{ho2020denoising, rombach2022high, ruiz2023dreambooth, nair2023steered, saharia2022photorealistic, ho2022cascaded, epstein2024diffusion, gu2022vector}. Introduced in \citet{ho2020denoising}, at its core, the diffusion models are built upon a Markov process and consists of two primary steps: (i) a forward process, in which noise is progressively added to the input data, and (ii) a denoising process, where a neural network is trained to iteratively remove noise. Popular architectures for this denoising process include U-Net \citep{ronneberger2015u} and transformer-based models \citep{vaswani2017attention}. \citet{nichol2021improved, ho2022classifier} demonstrate that these diffusion processes can be extended to model conditional distributions, wherein the generative process can be guided by external inputs, such as image labels. This can be achieved either through an explicit classifier or by training a single model that learns both unconditional and conditional distributions simultaneously. \citet{rombach2022high} further improve the computational efficiency of diffusion processes by proposing a method that operates in a lower-dimensional latent space instead of directly over image pixels. Specifically, images are first encoded into a latent space using a pre-trained variational autoencoder (VAE), diffusion is then applied in the latent space, and the generated output is mapped back to the image space through the VAE. For diffusion models with long Markov chains, inference can be time-consuming as the entire chain must be traversed. However, \citet{song2020denoising} show that the denoising process can be formulated as a non-Markovian chain, significantly reducing the number of steps required while avoiding the need for retraining the model. While Gaussian noise is most commonly used due to its smooth interaction with the continuous nature of images, \citet{hoogeboom2021argmax, austin2021structured} have also explored the extension of diffusion processes to model discrete distributions.

\vspace{0.5em}
\noindent\textbf{Diffusion for Structured Learning.} Although diffusion models have primarily been used for image generation, they have recently also been extended towards structured learning problems. \citet{chen2023diffusiondet} explore the application of diffusion models for object detection, formulating a diffusion process over bounding boxes, learning to denoise them by conditioning on an image, and generating corresponding labels as an auxiliary task. \citet{gu2024diffusioninst} extend this further to instance segmentation by incorporating an additional segmentation head. Diffusion models have also been applied in semantic segmentation, which focuses on generating accurate pixel-level labels for images. Methods in \citet{tan2022semantic, zbinden2023stochastic, wang2023segrefiner} define a diffusion process over pixel labels, learning to progressively denoise them to produce more precise segmentations. In all the aforementioned methods, diffusion is applied only to a single modality—whether bounding boxes or pixel labels—while other outputs, such as object categories or segmentations, are typically generated via auxiliary heads. In contrast, our proposed diffusion framework operates across multiple input modalities simultaneously (\eg boxes, labels, segmentation masks). Finally, while research in graph generation closely aligns with the task of scene graph generation, diffusion-based graph generation methods in \citet{jo2022score, vignac2022digress, yan2023swingnn} are unconditional in nature. This is in contrast to our proposed method that grounds graphs in a given image.

%% file: sections/2_background.tex
\section{Background}
\label{sec:formulation}

\subsection{Diffusion Models}
Diffusion models \citep{sohl2015deep, ho2020denoising} are probabilistic generative models that capture complex data distributions by learning to reverse a gradual noising process, generating samples via a sequence of iterative denoising steps. The diffusion framework is composed of two processes -- a \emph{forward diffusion process}, which progressively adds noise to data, and a \emph{reverse diffusion process}, wherein a model is trained to recover samples by removing the added noise step-by-step.

\noindent\textbf{Forward Process.} Let $\mathbf{x}_0 \sim q(\mathbf{x}0)$ denote a data point drawn from the empirical data distribution. The forward process defines a Markov chain $\mathbf{x}_{1:T} = \{\mathbf{x}_1, \mathbf{x}_2, ..., \mathbf{x}_T \}$ over $T$ discrete timesteps, where each random variable $\mathbf{x}_t \in \mathbb{R}^d$ represents a progressively corrupted version of the original data sample. The joint distribution of this forward process is defined as
\begin{align}
\label{eq:forwardprocess}
q\left(\mathbf{x}_{1:T} \mid \mathbf{x}_0\right)
:= \prod_{t=1}^{T} q\left(\mathbf{x}_t \mid \mathbf{x}_{t-1}\right),
\end{align}
where each transition distribution $q(\mathbf{x}_t \mid \mathbf{x}_{t-1})$ injects a small amount of noise into the previous state. The forward process is fixed and does not involve any learnable parameters.

\noindent\textbf{Reverse Process.}
The objective of the reverse process is to learn a set of parameters $\boldsymbol{\theta}$ that can effectively remove the corruption introduced by the forward process. For each step in the reverse process, the model estimates a conditional distribution $p_{\bm{\theta}}(\mathbf{x}_{t-1} \,|\, \mathbf{x}_t)$ to approximate the ground-truth denoising transition $q(\mathbf{x}_{t-1}\,|\,\mathbf{x}_t, \mathbf{x}_0)$. The training objective then is to minimize the distance between these two distributions, typically by using the Kullback-Leibler (KL) divergence.
\begin{align}
\label{eq:diffusiontraining}
\underset{\mathbf{\theta}}{\text{arg min}}\,D_{\text{KL}}\left(q\left(\mathbf{x}_{t-1}\,|\,\mathbf{x}_t, \mathbf{x}_0  \right) \, || \, p_{\bm{\theta}}\left(\mathbf{x}_{t-1}\,|\,\mathbf{x}_t\right)\right).
\end{align}

The type of noise added in each transition of the Markov chain described in Eq. (\ref{eq:forwardprocess}) can vary depending on the specific application and data characteristics. We will now outline the Gaussian diffusion framework, which is a widely used instantiation of diffusion models, and is particularly suitable for continuous data distributions such as images.

\subsection{Gaussian Diffusion} 
Gaussian diffusion models define a forward process that incrementally adds Gaussian noise to the data at each timestep. Formally, the forward transition at timestep $t$ is defined as,
\begin{align}
    \label{eq:gaussianforwardprocess}
    q\left(\mathbf{x}_t \,|\, \mathbf{x}_{t-1}\right) := \mathcal{N}(\mathbf{x}_t; \sqrt{1 - \beta_t} \mathbf{x}_{t-1}, \beta_t \mathbf{I}).
\end{align}
$\beta_t \in (0, 1)$ is a predefined variance schedule that controls the amount of noise added at each timestep $t$, where $\beta_t \rightarrow 1$ as $t \rightarrow T$. 

Over multiple steps, this process gradually corrupts the original data. When the magnitude of $\beta_t$ remains sufficiently small and the length of the Markov chain $T$ is large, the distribution of $\mathbf{x}_T$ converges to a standard Gaussian distribution $\mathcal{N}(0, \mathbf{I})$. 

A key property of the Gaussian diffusion forward process is that the marginal distribution of a noised $\mathbf{x}_t$ given data $\mathbf{x}_0$ admits a closed-form expression \citep{ho2020denoising},
\begin{align}
\label{eq:gaussianforwardprocessx0}
\begin{split}
    q\left(\mathbf{x}_t \, | \, \mathbf{x}_0\right) &:= \mathcal{N}\left(\mathbf{x}_t; \sqrt{\bar{\alpha}_t}\mathbf{x}_0,(1 - \bar{\alpha}_t)\mathbf{I}\right)\\
    \bar{\alpha}_t := &\prod_{i=0}^t\alpha_i;~~ \alpha_i := (1 - \beta_i)
\end{split}
\end{align}
This allows an arbitrary noised sample $\mathbf{x}_t$ to be efficiently obtained directly from $\mathbf{x}_0$, without the need to traverse through the Markov chain described in Eq. (\ref{eq:forwardprocess}). Such a formulation allows computation of the ground-truth denoising transition $q(\mathbf{x}_{t-1} | \mathbf{x}_t, \mathbf{x}_0)$ to also be formulated in a closed-form as,
\begin{align}
\begin{split}
    q\left(\mathbf{x}_{t-1} \,|\, \mathbf{x}_t, \mathbf{x}_0\right) &:= \mathcal{N}\left(\mathbf{x}_{t-1};\, \bm{\mu}_{q}(\mathbf{x}_t, \mathbf{x}_0), \bm{\Sigma}(t)\right) \\
    &:= \mathcal{N}\left(\mathbf{x}_{t-1};\,
\underbrace{\frac{\sqrt{\alpha_t}(1-\bar{\alpha}_{t-1})\,\mathbf{x}_t
      +\sqrt{\bar{\alpha}_{t-1}}(1-\alpha_t)\,\mathbf{x}_0}
     {1-\bar{\alpha}_t}}_{\bm{\mu}(\mathbf{x}_t, \mathbf{x}_0)},\;
\underbrace{\frac{(1-\alpha_t)(1-\bar{\alpha}_{t-1})}{1-\bar{\alpha}_t}\,\mathbf{I}}_{\bm{\Sigma}(t)}
\right)
\end{split}
\end{align}

To approximate the aforementioned posterior transition as closely as possible, the learned denoising step $p_{\bm{\theta}}(\mathbf{x}_{t-1} \,|\, \mathbf{x}_t)$ is also assumed to be Gaussian,
\begin{align}
\begin{split}
    \label{eq:ptheta}
    p_{\bm{\theta}}(\mathbf{x}_{t-1} \,|\, \mathbf{x}_t) &:= \mathcal{N}(\mathbf{x}_{t-1}; \bm{\mu}_{\bm{\theta}}(\mathbf{x}_t, t), \bm{\Sigma}(t))\\
    &:= \mathcal{N}\left(\mathbf{x}_{t-1};\,
\underbrace{\frac{\sqrt{\alpha_t}(1-\bar{\alpha}_{t-1})\,\mathbf{x}_t
      +\sqrt{\bar{\alpha}_{t-1}}(1-\alpha_t)\,\hat{\mathbf{x}}_\theta(\mathbf{x}_t, t)}
     {1-\bar{\alpha}_t}}_{\bm{\mu}_{\bm{\theta}}(\mathbf{x}_t, t)},\;
\underbrace{\frac{(1-\alpha_t)(1-\bar{\alpha}_{t-1})}{1-\bar{\alpha}_t}\,\mathbf{I}}_{\bm{\Sigma}(t)}
\right)
\end{split}
\end{align}
where $\bm{\mu}_{\bm{\theta}}$ is the mean function parameterized by a neural network conditioned on both the noisy input $\mathbf{x}_t$ and the timestep $t$. In practice, this mean function is often implemented to predict the denoised sample $\hat{\mathbf{x}}_\theta(\mathbf{x}_t, t)$, which serves as an estimate of the original sample $\mathbf{x}_0$. Additionally, note that the covariance is assumed to be identical to the true posterior transition, which simplifies the learning problem to matching the mean of $p_{\bm{\theta}}$ to that of $q$.

As shown in \citet{ho2020denoising}, by choosing an appropriate parameterization of the mean function $\bm{\mu}_{\bm{\theta}}$, the KL divergence objective simplifies in Eq. (\ref{eq:diffusiontraining}) to a mean squared error loss,
\begin{align}
    \label{eq:loss}
    \underset{\mathbf{\theta}}{\text{arg min}} \left|\left|\hat{\mathbf{x}}_\theta(\mathbf{x}_t, t) - \mathbf{x}_0 \right|\right|^2_2
\end{align}
Once trained, diffusion models generate new samples by iteratively applying the reverse diffusion process starting from pure noise. Specifically, sampling begins by drawing an initial latent $\mathbf{x}_T \sim \mathcal{N}(0, \mathbf{I})$, and then repeatedly applying the learned reverse transitions $p_{\bm{\theta}}(\mathbf{x}_{t-1} \mid \mathbf{x}_t)$, as defined in Eq. (\ref{eq:ptheta}), for $t = T, T-1, \dots, 1$. 

\noindent\textbf{Conditioning.} Diffusion models can be extended to learn conditional distributions $p_{\bm{\theta}}(\mathbf{x}_{t-1} \mid \mathbf{x}_t, \mathbf{c})$, where $\mathbf{c}$ denotes auxiliary information. This conditioning may represent class labels, text embeddings, or other semantic signal that guides generation. This information is typically passed as an additional input to the denoising network, for example through feature modulation or cross-attention. The overall training objective is analogous to unconditional case,
\begin{align}
    \label{eq:conditioningloss}
    \underset{\mathbf{\theta}}{\text{arg min}} \left|\left|\hat{\mathbf{x}}_\theta(\mathbf{x}_t, t, \mathbf{c}) - \mathbf{x}_0 \right|\right|^2_2
\end{align}

%% file: sections/3_approach.tex
\section{Approach}
\label{sec:approach}
\begin{figure}[t]
    \centering
    \includegraphics[width=0.98\textwidth]{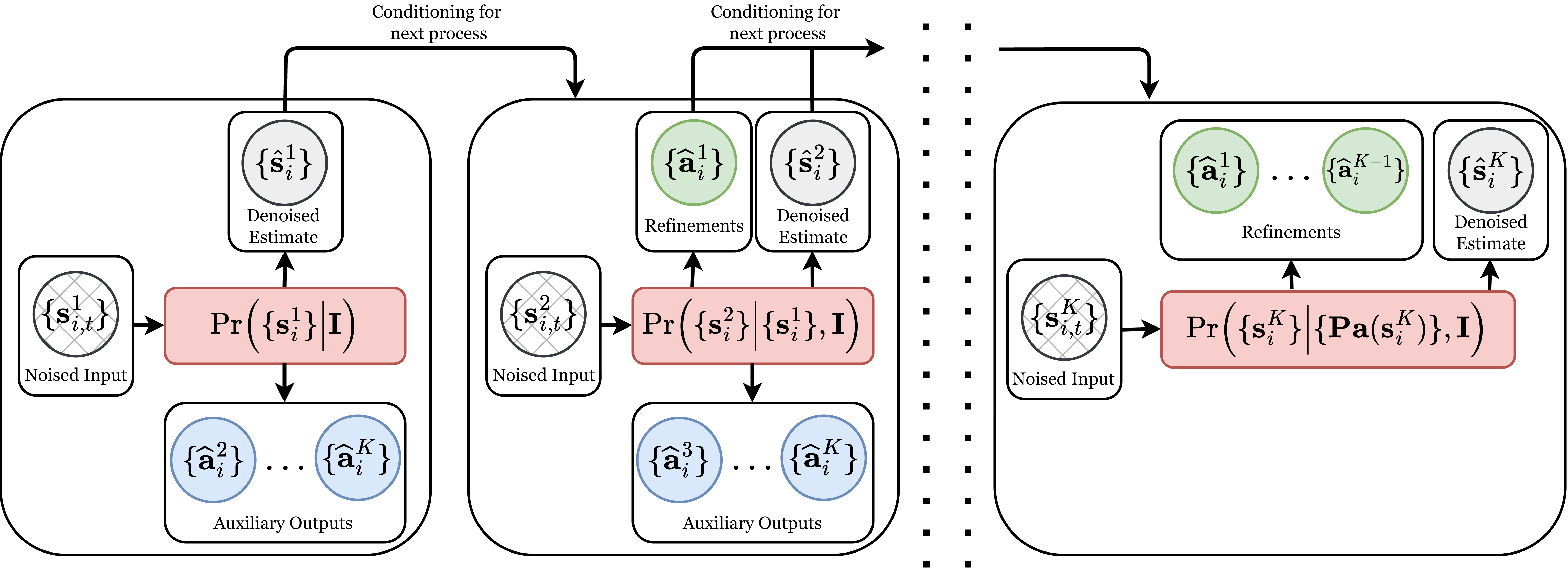}
    \caption{\textbf{Modular Diffusion Models.} We factorize the joint distribution over structured outputs into a sequence of conditional distributions, each modeled by its own diffusion process (\textcolor{Salmon}{red}). At each stage, a module takes a noisy input along with relevant conditioning, and produces a denoised estimate, updated conditioning (\textcolor{YellowGreen}{green}), and auxiliary outputs (\textcolor{NavyBlue}{blue}).}
    \label{fig:mdm-overview}
\end{figure}

In this section, we first outline the general MDM framework. We then present specific instantiations of this framework on three structured learning tasks -- object detection, instance segmentation, and scene graph generation.

\subsection{Modular Diffusion Models}
\label{sec:mdm}
Let $\mathbf{S} = \{\mathbf{S}_i\}_{i\leq n}$ represent the set of structured representations for a specific image $\mathbf{I}$. 
Each structured representation can be further decomposed into distinct components $\mathbf{S}_i = (\mathbf{s}_i^1, \dots, \mathbf{s}_i^K)$. As an example, for the task of instance segmentation, each $\mathbf{S}_i$ corresponds to an individual object instance and is represented as $\mathbf{S}_i = \langle \mathbf{b}_i, \mathbf{l}_i, \mathbf{m}_i \rangle$, with $\mathbf{b}_i$, $\mathbf{l}_i$, and $\mathbf{m}_i$ denoting the bounding box, class label, and segmentation mask respectively. Therefore, the underlying learning problem can be formulated as estimating the joint distribution $\Pr(\,\mathbf{S} \,|\,\mathbf{I})$. Directly modeling this joint distribution through a diffusion process, however, can present significant challenges, particularly as the output space grows combinatorially with the increasing cardinality of $\mathbf{S}$. Furthermore, the components $\mathbf{s}_i$ may exhibit complex interdependencies and have distinct data types (\eg boxes, labels), adding an additional layer of difficulty in directly capturing the joint distribution. Modeling these complex, heterogeneous dependencies efficiently therefore requires a more structured formulation -- one that factorizes the joint distribution into simpler, easier-to-learn conditional distributions.

In particular, for a given structured representation $\mathbf{S}$, one can define a plausible factorization via a Directed Acyclic Graph (DAG) $\mathbf{G}_\mathbf{S} = (\mathbf{S}, \mathbf{E}_\mathbf{S})$, where $\mathbf{E}_\mathbf{S}$ represents the directed edges denoting the conditional dependencies between the components $\mathbf{s}_i$. Under this DAG structure, the joint distribution $\Pr(\mathbf{S}\,|\,\mathbf{I})$ can be factorized as,
\begin{align}
    \label{eq:mdm-factorization}
    \Pr\Bigl(\,\mathbf{S} \,\Big|\, \mathbf{I}\,\Bigr) = \prod_{i=1}^{K} \Pr\Bigl(\, \{ \mathbf{s}^k_i \} \,\Big|\, \{\mathbf{Pa}( \mathbf{s}^k_i)\}, \mathbf{I} \, \Bigr)    
\end{align}
where $\mathbf{Pa}(\mathbf{s}^k_i)$ denotes the set of parents of $\mathbf{s}^k_i$ in the DAG. Note that most existing methods for structured learning tasks often implicitly incorporate such a factorization via model architecture design.

Our proposed MDM framework leverages this factorization under a DAG structure to make jointly learning a diffusion process easier. Specifically, MDM \emph{independently} models each conditional $\Pr(\{\mathbf{s}_i\} \,|\, \{\mathbf{Pa}(\mathbf{s}_i)\},\, \mathbf{I})$ as a separate diffusion model. Since each model is now responsible for learning the distribution over a single component $\mathbf{s}^k_i$, this approach reduces the complexity of the learning task. This modularity enables each diffusion process to be tailored to its corresponding component, allowing for flexibility in the choice of diffusion process and architectural design. Additionally, as each model now operates over a smaller output space, learning is more tractable. 

During inference, a valid structured representation can easily be obtained without any additional training, via ancestral sampling over the specified DAG. For each component $\mathbf{s}^k_i$, this involves taking multiple denoising steps starting from random noise. Since each individual diffusion model is trained independently, it can be run for a varying number of denoising steps, allowing for fine-grained control over the generation process. As such, this flexibility in the number of denoising steps for each model leads to MDMs being able to generate exponentially large number of potential structured representations while only requiring linear number of diffusion processes.

\subsection{MDM for Detection and Instance Segmentation}
\label{sec:mdm-seg}
\begin{figure}[t]
    \centering
    \includegraphics[width=0.98\textwidth]{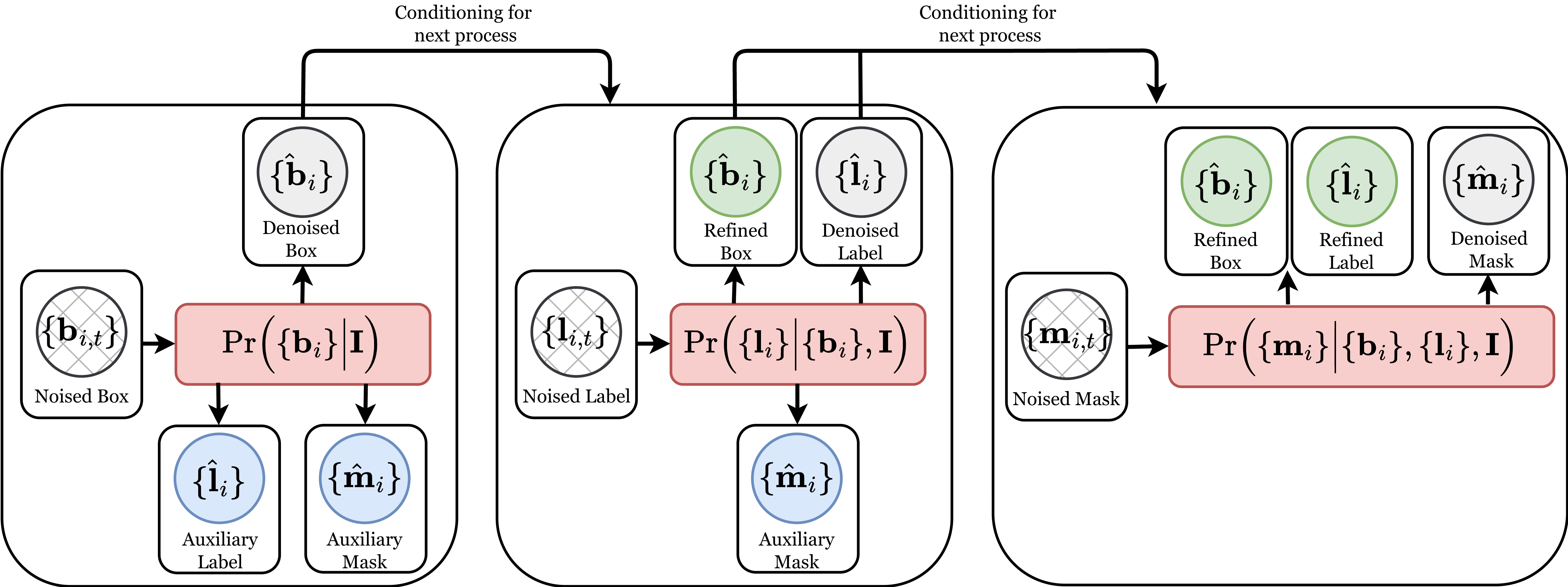}
    \caption{\textbf{Modular Diffusion Model for Instance Segmentation.} This consists of three separate diffusion processes - (i) box diffusion, which denoises bounding boxes conditioned on image features while producing auxiliary label and mask predictions; (ii) label diffusion, which denoises class labels conditioned on the current box estimates and image features, while refining the box and generating auxiliary outputs; and (iii) mask diffusion, which denoises segmentation masks conditioned on the estimated boxes, labels, and image features, while refining both box and label predictions.}
    \label{fig:mdm-seg}
\end{figure}
In this section, we present a specific instantiation of our proposed MDM framework on the task of instance segmentation, which requires models to accurately localize, label, and outline the boundaries of each distinct object instance within an image. Note that, within our framework, the task of object detection can be viewed as a special case of instance segmentation in which the mask component is omitted. 

For an input image $\mathbf{I}$, the instance segmentation task outputs a set of representations 
$\mathbf{S}^{\text{seg}} = \{(\mathbf{b}_i, \mathbf{l}_i, \mathbf{m}_i)\}$, which typically consists of three components -- a bounding box $\mathbf{b}_i$ that provides a spatial location for each instance, a distribution over class labels $\mathbf{l}_i$ that specifies the likelihood of the instance belonging to each object category, and a binary mask $\mathbf{m}_i$ that defines a pixel-wise segmentation for the object. 

For the task of instance segmentation, our proposed MDM framework factorizes the joint distribution $\Pr(\mathbf{S}^{\text{seg}}\,|\,\mathbf{I})$ using the dependency structure specified by the DAG,

\begin{align}
    \label{eq:segfactorization}
    \begin{split}
    \tikzmarknode[]{a}{\mathbf{b}} \;\;\;\;\;\;\;\;\; & \tikzmarknode[]{b}{\mathbf{l}} \;\;\;\;\;\;\;\;\; \tikzmarknode[]{c}{\mathbf{m}} \\
    \Pr\Bigl(\mathbf{S}^{\mathrm{seg}} \,\Big|\, \mathbf{I}\Bigr)
= \Pr\Bigl(\{\mathbf{b}_i\} \,\Big|\, \mathbf{I}\Bigr) 
\; \cdot \; \Pr&\Bigl(\{\mathbf{l}_i\} \,\Big|\, \{\mathbf{b}_i\}, \mathbf{I}\Bigr) 
\; \cdot \; \Pr\Bigl(\{\mathbf{m}_i\} \,\Big|\, \{(\mathbf{b}_i, \mathbf{l}_i)\}, \mathbf{I}\Bigr)
    \end{split}
\end{align}
\begin{tikzpicture}[remember picture,overlay]
\draw[-latex]([yshift=.7ex, xshift=1.0ex]a.north) to[bend left]node[above]{} ([yshift=.7ex, xshift=-1ex]c.north);
\draw[-latex]([xshift=2.0ex]a.west) to[right]node[right]{} ([xshift=-2.0ex]b.east);
\draw[-latex]([xshift=2.0ex]b.west) to[right]node[right]{} ([xshift=-3.0ex, yshift=0.3ex]c.east);
\end{tikzpicture}

This factorization follows the natural structure of the instance segmentation task. The model first predicts the locations of candidate objects in the image using bounding boxes $\{\mathbf{b}_i\}$. Given these locations, it then predicts the semantic category distribution $\{\mathbf{l}_i\}$ for each detected instance. Finally, conditioned on the image, the predicted box, and the corresponding class label distribution, the model generates the instance mask $\{\mathbf{m}_i\}$, which captures the detailed pixel-level shape of the object.

We define three separate diffusion processes, each corresponding to one of the conditional distributions in Eq. (\ref{eq:segfactorization}). The forward process for each of these conditionals is described below.

\textbf{Box Diffusion.} Each bounding box is parameterized by the continuous tuple $(x, y, w, h)$, representing the box center coordinates, and its width and height. Gaussian diffusion can therefore be applied directly by perturbing these parameters at timestep $t$ using the standard forward process (Eq. \ref{eq:gaussianforwardprocessx0}).

\textbf{Label Diffusion.} The label associated with each box is a categorical variable in $[1, C]$, where $C$ denotes the number of classes. Although diffusion can be defined directly over discrete variables \citep{austin2021structured}, we instead adopt a simpler Gaussian formulation using a continuous relaxation of discrete labels \citep{dieleman2022continuous, chen2022analog, chen2023generalist}. Specifically, each label is represented as a one-hot vector $\mathbf{l}_i = [0, \dots, 1, \dots]$, where the $j$-th component $\mathbf{l}_i[j] = 1$.

\textbf{Mask Diffusion.} Traditionally, diffusion over masks is treated as a semantic segmentation problem, where each pixel is associated with a categorical label \citep{zbinden2023stochastic}. In our MDM framework, as labels are modeled separately, we treat masks as binary spatial masks $\in [0, 1]^{W \times H}$ corresponding to individual object instances. Similar to label diffusion, we assume a continuous relaxation by downsampling masks to a smaller resolution $[0, 1]^{W/\tau \times H/\tau}$, where $\tau$ is the reduction factor. This has two benefits -- First, the relaxed mask representation allows Gaussian diffusion to be applied directly. Second, performing diffusion at a reduced spatial resolution significantly lowers computational cost, as the diffusion process operates on a smaller spatial grid. This is reasonable in practice, as instance segmentation methods often predict masks at relatively coarse resolutions before projecting them back to the original image size \citep{he2017mask}.

\subsection{MDM for Scene Graph Generation}
\label{sec:mdm-sg}

Scene graphs are a graphical representation wherein the nodes correspond to the entities present in the scene, with edges denoting the relationships between pairs of nodes. Each node is grounded within the image through its spatial location, detailed by a bounding box, and the corresponding class label. This graph is often represented using a set of \texttt{<subject, predicate, object>} triplets, where \texttt{subject} and \texttt{object} are nodes in the graph, and \texttt{predicate} is the relationship between them. 

Therefore, the task of scene graph generation can be viewed as learning a set of the aforementioned triplets, and the corresponding representation $\mathbf{S}^{\text{sg}} = \{(\mathbf{b}_i^s, \mathbf{b}_i^o, \mathbf{l}_i^s, \mathbf{l}_i^o, \mathbf{l}_i^p)\}$ that is expected as output consists of five components - bounding boxes $\mathbf{b}_i^s$ and $\mathbf{b}_i^o$ that localize the subject and object entities, category labels $\mathbf{l}_i^s$ and $\mathbf{l}_i^o$ for the corresponding boxes, and a relationship label $\mathbf{l}_i^p$ defining the interaction between them.

We define the following dependency structure to instantiate our proposed MDM framework for the task of scene graph generation.
\begin{align}
    \label{eq:scenegraphfactorization}
    \begin{split}
    \tikzmarknode[]{x}{[\mathbf{b}^s, \mathbf{b}^o]} \;\;\;\;&\;\;\;\;\;  \tikzmarknode[]{y}{[\mathbf{l}^s, \mathbf{l}^o, \mathbf{l}^p]} \\
    \Pr\Bigl(\mathbf{S}^{\mathrm{sg}} \,\Big|\, \mathbf{I}\Bigr) 
= \Pr\Bigl(\{(\mathbf{b}_i^s, \mathbf{b}_i^o)\} \,\Big|\, \mathbf{I}\Bigr)  \; &\cdot \; \Pr\Bigl(\{(\mathbf{l}_i^s, \mathbf{l}_i^o, \mathbf{l}_i^p)\} \,\Big|\, \{(\mathbf{b}_i^s, \mathbf{b}_i^o)\}, \mathbf{I}\Bigr)
    \end{split}
\end{align}
\begin{tikzpicture}[remember picture,overlay]
\draw[-latex]([xshift=7.5ex]x.west) to[right]node[right]{} ([xshift=-9.0ex]y.east);
\end{tikzpicture}

Intuitively, the aformentioned factorization first generates candidate of bounding boxes that are likely to exhibit an interaction. Given these box pairs, the model then generates the corresponding object categories and the relationship label.

We therefore define two separate diffusion processes, each corresponding to one of the conditional distributions in Eq. (\ref{eq:scenegraphfactorization}). The forward process for each of these conditionals is described below.

\textbf{Proposal Diffusion.} Scene graph grounds each relation to the image by a pair of bounding boxes. Therefore, the input representation for the proposal diffusion then is the concatenated continuous vector $[\mathbf{b}_i^s, \mathbf{b}_i^o]$, where each box $\mathbf{b}_i^e$ is parameterized by the continuous tuple $(x, y, w, h)$, representing the box center coordinates and its width and height. As these parameters lie in a continuous space, Gaussian diffusion can be applied directly.

\textbf{Triplet Diffusion.} The subject, object label, and predicate labels are categorical variables defined over the entity and relationship vocabularies. Similar to the label diffusion process described for instance segmentation (Section \ref{sec:mdm-seg}), we adopt a continuous relaxation for these discrete variables. The input to triplet diffusion is therefore the concatenated continuous vector $[\mathbf{l}_i^s, \mathbf{l}_i^o, \mathbf{l}_i^p]$, where each component is a one-hot vector corresponding to its respective category. Gaussian noise is then applied to this continuous representation during the forward diffusion process.


\subsection{Transformer Based Denoising Process}
\label{sec:tranformer-denoising}
\begin{figure}[t]
    \centering
    \includegraphics[width=0.98\textwidth]{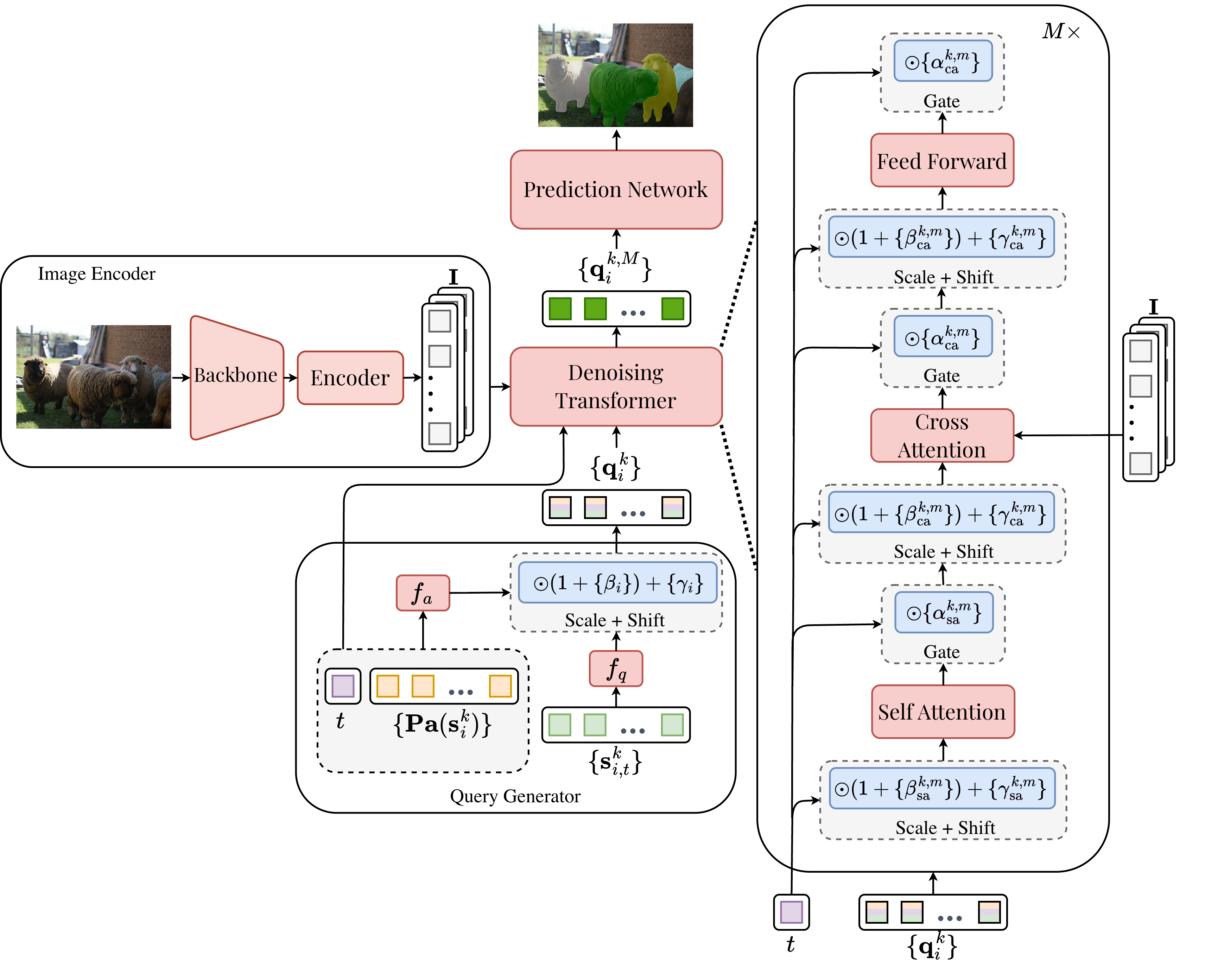}
    \caption{\textbf{Transformer-based Architecture.} A simple transformer-based instantiation of the proposed MDM framework that consists of -- (i) an image encoder that extracts multi-scale features from the input image, (ii) a query generation module that maps noisy inputs into queries for the denoising transformer, (iii) a stack of $M$ time-conditioned denoising decoder layers that iteratively refine representations through attention and feed-forward operations, and (iv) a prediction head that produces the final structured outputs.}
    \label{fig:mdm-architecture}
\end{figure}

In this section we present a concrete instantiation of the reverse diffusion process for the MDM framework. Specifically, we modify the transformer based architecture in \citet{li2023mask} to operate within MDM framework, as highlighted in Figure \ref{fig:mdm-architecture}. The choice of a transformer-based architecture is motivated by their ability to effectively model set-based structured outputs and capture complex interactions between elements through attention mechanisms. Importantly, it should be noted that the MDM framework itself is architecture agnostic and can be implemented using a variety of design choices. 

Recall that our proposed MDM framework defines a diffusion process for each conditional $\Pr(\{\mathbf{s}_i^k\} \,|\, \{\mathbf{Pa}(\mathbf{s}_i^k)\}, \mathbf{I})$ in Eq. \ref{eq:mdm-factorization}. As such, the reverse process is implemented via a multi-layer transformer-based decoder, which denoises a noisy input at timestep $t$, ${\mathbf{s}}_{i,t}^k$, conditioned on the image features $\mathbf{I}$ and any ancestral input $\mathbf{Pa}(\mathbf{s}_i^k)$. For example, in the case of the mask diffusion described in Section \ref{sec:mdm-seg}, these ancestral inputs would include the denoised box $\{\hat{\mathbf{b}}_i\}$ and label estimates $\{\hat{\mathbf{l}}_i\}$ from the previous diffusion processes. Additionally, it should be noted that the image features $\mathbf{I}$ are obtained by passing the raw image through a combination of convolutional backbone and a multi-layer transformer encoder, akin to methods in \citet{cheng2022masked,li2023mask}.

We now describe the key components of this architecture and explain how they are adapted to serve as the denoising network for the reverse diffusion process.

\subsubsection{Image Encoder}
\label{sec:image-encoder}
Following \citet{li2023mask}, for each image $\mathbf{I}$, the encoder first employs a deep convolutional network (\eg ResNet \citep{he2016deep}) to extract multi-scale spatial feature maps $[\mathbb{R}^{c \times W_1 \times H_1}, \dots, \mathbb{R}^{c \times W_r \times H_r}]$ where $c$ is the number of channels, and $(W_r, H_r)$ denotes the spatial resolution at each scale. These feature maps are concatenated and passed through a multi-layer transformer encoder, which transforms them into position-aware flattened image representations $\mathbf{I} \in \mathbb{R}^{c \times \sum_r W_rH_r}$. These are subsequently used in the decoder to condition the denoising process.

\subsubsection{Denoising Decoder}
\label{sec:denoising-decoder}

The denoising decoder is tasked with recovering a set of clean estimates $\{\hat{\mathbf{s}}^k_{i}\}$ from the noisy input set $\{{\mathbf{s}}^k_{i, t}\}$ at a given timestep $t$, and is implemented as a multi-layer transformer decoder. Unlike traditional transformer-based methods (\eg \citet{carion2020end}) that use learnable queries to initialize the decoding process, we formulate the input queries $\{\mathbf{q}_i^k\}$ as a function of the noisy input, modulated by time-aware ancestral conditioning. Specifically,
\begin{align}
    \begin{split}
        \label{eq:input-projection}
        \boldsymbol{\beta}_i^k, \boldsymbol{\gamma}_i^k &= f_a\left(\mathbf{Pa}(\mathbf{s}_i^k),\,t\right), \\
        \mathbf{q}_i^k &= f_q\left({\mathbf{s}}^k_{i, t}\right) \odot \left(1 + \boldsymbol{\beta}_i^k\right) + \boldsymbol{\gamma}_i^k
    \end{split}
\end{align}
where $f_a, f_q$ are neural networks, and $\odot$ refers to elementwise multiplication. This formulation enables the queries to dynamically adapt to the noisy inputs as well as any additional contextual information, conditioned on the current diffusion timestep.

The transformer decoder, consisting of $M$ layers, takes as input the set of queries $\{\mathbf{q}_i^k\}$ and iteratively denoises them. Each layer applies self-attention (sa) on the queries, cross-attention (ca) to the image features $\mathbf{I}$, and a feed-forward network (ffn). To incorporate timestep information, we adopt the zero-initialized adaptive layer normalization (adaLN-Zero) \citep{peebles2023scalable}, which modulates each update with time embeddings. 

For layer $m \le M$, the update can be written as a sequence of operations,
\begin{align}
\begin{split}
\label{eq:time-projection}
\mathbf{q}_{i,\mathrm{sa}}^{k,m}
&= \mathcal{F}_{\mathrm{sa}}^{m}\bigl(\mathbf{q}_{i}^{k,m-1}, t\bigr), \\
\mathbf{q}_{i,\mathrm{ca}}^{k,m}
&= \mathcal{F}_{\mathrm{ca}}^{m}\bigl(\mathbf{q}_{i,\mathrm{sa}}^{k,m}, \mathbf{I}, t\bigr), \\
\mathbf{q}_{i}^{k,m}
&= \mathcal{F}_{\mathrm{ffn}}^{m}\bigl(\mathbf{q}_{i,\mathrm{ca}}^{k,m}, t\bigr),
\end{split}
\end{align}

where each transformation $\mathcal{F}_{x }^{m};\; x \in \{\text{sa}, \text{ca}, \text{ffn}\}$ can be expressed as,
\begin{align}
    \mathcal{F}_{x}^{m}(\mathbf{q}, \mathbf{c}, t)
    =
    \mathbf{q}
    +
    \boldsymbol{\alpha}_{x}^{k,m}
    \odot
    L_{x,m}\!\Big(
        \mathbf{q}
        \odot (1 + \boldsymbol{\beta}_{x}^{k,m})
        +
        \boldsymbol{\gamma}_{x}^{k,m},
        \mathbf{c}
    \Big)
\end{align}
where $L_{x,m}(\cdot)$ is the transformer module (self-attention, cross-attention, or feed-forward network). Additionally, conditioning input $\mathbf{c} = \varnothing$ for $x \in {\text{sa}, \text{ffn}}$, and $\mathbf{c} = \mathbf{I}$ for $x = \text{ca}$. The time-dependent modulation parameters $(\boldsymbol{\alpha}, \boldsymbol{\beta}, \boldsymbol{\gamma})$ are computed as,
\begin{align}
    (\boldsymbol{\alpha}^{k,m}_x, \boldsymbol{\beta}^{k,m}_x, \boldsymbol{\gamma}^{k,m}_x) &= f_{x,m}(t)
\end{align}
where $f_{x,m}(t)$ is a learned neural network. After $M$ iterations of denoising, the set of clean estimate $\{\hat{\mathbf{s}}^k_{i}\}$ are obtained by passing the output of the last layer $\{\mathbf{q}_{i}^{k,M}\}$ through a prediction network.

\subsection{Training}
\label{sec:training}
Each conditional diffusion process $\Pr(\{\mathbf{s}_i^k\} \,|\, \{\mathbf{Pa}(\mathbf{s}_i^k)\}, \mathbf{I})$ can be trained in an end-to-end fashion. Note that, following transformer-based set prediction frameworks \citep{carion2020end,zhu2020deformable}, we use a fixed number of queries $Q_{\text{train}}$ during training. As the number of ground-truth elements is typically significantly smaller than $Q_{\text{train}}$, the ground-truth set is padded with background (no-object) entries. For instance, in instance segmentation, this padding consists of random boxes, a background label, and zero-valued segmentation masks. The noisy input $\{\mathbf{s}^k_{i,t}\}$ for each diffusion process is then computed over this padded set.

As the denoised estimates $\{\hat{\mathbf{s}}^k_{i}\}$ obtained from a diffusion process form an unordered set, we adopt a bipartite matching algorithm \citep{carion2020end} to align them with the ground truth set $\{\mathbf{s}^k_{i}\}$. However, as most structured predicted tasks involves correlated components, computing this matching independently for each element can be suboptimal. For example. matching based on boxes alone can be ambiguous when multiple boxes are close or overlapping, potentially assigning the wrong ground-truth object and producing inconsistent conditioning for subsequent labels and mask diffusion processes. To mitigate this, we additionally compute estimates for the other components of the structured output, $\{\hat{\mathbf{a}}_i^{v}\}, v \neq k$, by passing the output queries  $\{\mathbf{q}_{i}^{k,M}\}$ through the corresponding prediction networks. These estimates can be divided into two sets -- (i) refinements over previous predictions $v \in \mathbf{Pa}(\mathbf{s}_i^k)$, and (ii) auxiliary outputs $v \notin \mathbf{Pa}(\mathbf{s}_i^k)$. It should be noted that the auxiliary outputs are only computed to guide model training and are \emph{not} passed as conditioning to subsequent diffusion processes. 

We define the joint matching cost $\mathcal{L}_{\sigma}$ for a given matching $\sigma$ as
\begin{align}
\mathcal{L}_{\sigma} =
\left[
\sum_i \mathcal{L}^{k}\!\left(\hat{\mathbf{s}}_{i}^{k}, \mathbf{s}^{k}_{\sigma(i)}\right)
+ \sum_i \mathcal{L}^{\text{diff}}\!\left(\hat{\mathbf{s}}_{i}^{k}, \mathbf{s}^{k}_{\sigma(i)}\right)
+ \sum_{v \neq k} \sum_i \mathcal{L}^{v}\!\left(\hat{\mathbf{a}}_{i}^{v}, \mathbf{s}^{v}_{\sigma(i)}\right)
\right].
\end{align}

where $\mathcal{L}^{k}$ is the function that defines the loss for component $k$ (\eg combination of L-1 and generalized IoU loss for the boxes), and $\mathcal{L}^{\text{diff}}$ is the diffusion loss (Eq. \ref{eq:conditioningloss} in case of Gaussian diffusion). A matching algorithm computes optimal permutation $\widehat{\sigma}$ as,
\begin{align}
\widehat{\sigma} = \arg\min_{\sigma}\, \mathcal{L}_{\sigma}.
\end{align}

The choice of matching algorithm is task-dependent. For instance, in object detection and instance segmentation, the Hungarian algorithm is used~\citep{carion2020end}, whereas for scene graph generation a multi-assignment heuristic can be employed~\citep{kim2024groupwise}.

Finally, to improve convergence, we apply the loss $\mathcal{L}_{\widehat\sigma}$ to the outputs from every intermediate layer, as is the case in \citet{carion2020end,li2023mask,cheng2022masked}. Specifically, for each intermediate decoder layer $m < M$, we compute the corresponding denoised and auxiliary outputs from $\{\mathbf{q}_{i}^{k,m}\}$, determine an optimal matching, and apply the joint matching loss.

Our proposed MDM framework allows each diffusion network corresponding to a conditional distribution (Eq. \ref{eq:mdm-factorization}) to be trained independently. In practice, however, this would require maintaining separate parameters for both the image encoders and denoising decoders for each task, which can be impractical under memory constraints.
To address this, we share the encoder and denoising decoder weights across all tasks to reduce memory overhead. Task-specific behavior is induced through projection parameters $f_q$ (Eq. \ref{eq:input-projection}) and $f_{x,m}$ (Eq. \ref{eq:time-projection}), which modulate the query representations. This design enables the shared parameters to operate across tasks while still allowing each conditional diffusion process to retain task-specific conditioning. An additional benefit of this parameter sharing is improved training and inference speeds as the encoder needs to compute image features $\mathbf{I}$ only once per batch, which can then be reused across tasks during both training and inference.

\subsection{Inference}
\label{sec:inference}

As described in Section \ref{sec:mdm}, inference in the proposed MDM framework is performed via ancestral sampling over the underlying DAG. However, as noted in \citet{song2020denoising}, naively traversing the full Markov chain during sampling can be computationally expensive when the number of diffusion timesteps $T$ is large. As such, we adopt the DDIM sampling procedure \citep{song2020denoising}, which accelerates inference by transforming the original diffusion process into a non-Markovian process, allowing sampling with substantially fewer timesteps.

Furthermore, since each conditional distribution in the DAG is modeled as a separate diffusion process, the number of sampling steps can be chosen separately for each component. For example, in an instance segmentation model trained using the proposed MDM framework, the box diffusion process can be run for $T^{\text{box}}$ steps, the label diffusion can be run for $T^{\text{label}}$ steps, and the mask diffusion can be run for $T^{\text{mask}}$ steps, enabling fine-grained control over the generation process.

%% file: sections/4_experiments.tex
\section{Experiments}
We evaluate the proposed MDM framework on three structured prediction tasks: object detection, instance segmentation, and scene graph generation. While MDM achieves competitive performance across all tasks, our primary objective is to emphasize two key points: (i) the framework’s generality, demonstrating its applicability across a diverse set of structured prediction problems, and (ii) the effectiveness of a simple transformer-based instantiation, which attains strong results without relying on task-specific architectural components, data augmentation, or optimization techniques.

\subsection{Object Detection and Instance Segmentation}
\label{sec:mscoco-results}
\input{tables/coco-instance-segmentation}
\input{tables/coco-obj-detection}

\textbf{Dataset.} We evaluate performance on the MS-COCO 2017 dataset \citep{lin2014microsoft}. MS-COCO contains approximately $160$K images across $80$ object categories, split into $118$K training images, $5$K validation images, and $41$K test images. Since ground-truth annotations for the test set are not publicly available, following prior work \citep{li2022dn}, we report results on the validation set. Performance is measured using average precision (AP) across multiple IoU thresholds and object scales.

\textbf{Implementation Details.} We train the transformer-based framework described in Section \ref{sec:tranformer-denoising} on the task of object detection using the MDM formulation in Section \ref{sec:mdm-seg}. For fair comparison with prior methods, we adopt a ResNet-50 backbone \citep{he2016deep} as the feature extractor in the image encoder. The encoder comprises $6$ transformer layers, while the denoising decoders use $6$ layers for object detection and $9$ layers for instance segmentation. As discussed in Section~\ref{sec:training}, parameters are shared across all denoising decoders, resulting in a model with a comparable number of parameters to the baseline methods. Finally, we set $Q_{\text{train}} = 300$ and, following the baseline in \citep{chen2023diffusiondet}, employ a box-renewal strategy \citep{chen2023diffusiondet} during box diffusion inference to better align the inference procedure with training.

\textbf{Results -- Instance Segmentation.} We evaluate MDM for instance segmentation on MS-COCO in Table \ref{tab:coco-inst-results}. Our approach demonstrates a significant performance advantage over existing diffusion-based methods \circled{6}-\circled{7} and achieves parity with state-of-the-art deterministic segmentors \circled{8}. Specifically, MDM \circled{13} outperforms DiffusionInst \citep{gu2024diffusioninst} \circled{7} -- which only applies diffusion to bounding boxes -- by $7.4$ AP$^\text{mask}$ using the same number of box refinement steps ($T^{\text{box}}=4$). By further incorporating label and mask diffusions, MDM extends this margin to $8.4$ AP$^\text{mask}$ \circled{13}, highlighting the effectiveness of our factorized approach.

When compared to determinstic methods, MDM achieves performance on-par with the strong baseline in Mask DINO \citep{li2023mask} \circled{5}. Beyond raw metrics, the generative nature of MDM offers a distinct advantage over deterministic models -- Mask DINO is limited to a single deterministic point-estimate prediction. In contrast, MDM attempts to model the underlying data distribution, enabling it to sample multiple plausible outputs for the same input image. This allows our model to handle uncertainty and resolve ambiguous cases in complex scenes that a single-pass deterministic model might overlook. A qualitative analysis of this behavior is provided in Section \ref{sec:qualitative-analysis}.

\textbf{Results -- Object Detection.} We contrast MDM against existing detection methods in Table \ref{tab:coco-det-results}. MDM \squared{16} - \squared{20} consistently outperforms most deterministic baselines \squared{1} – \squared{12} and all diffusion-based counterparts \squared{14} – \squared{15} across all metrics. Additionally, we achieve performance on-par with the competitive baseline in DINO-DETR \citep{zhang2022dino} (\squared{13}), which represents the most comparable baseline to our specific transformer-based implementation (Section \ref{sec:tranformer-denoising}). Additionally, MDM benefits from a generative formulation (as discussed in the instance segmentation setting), enabling it to model the underlying data distribution and produce multiple plausible predictions.

Another mode of comparison is the number of object queries used at inference. Deterministic baselines operate under a fixed query budget (typically $300$ queries). Recently, several high-performing variants \citep{liu2022dab,li2022dn,zhang2022dino} \squared{11}–\squared{13} report results using an effective query count of $900$. Although increasing the number of queries generally improves detection performance, this typically requires retraining, which is computationally expensive and time-consuming. In contrast, our proposed MDM framework can scale to an arbitrary number of queries at inference without any retraining. As shown in \squared{22}, a model trained with $300$ queries can be directly evaluated in the $900$-query regime, yielding consistent performance without additional optimization. This provides a more flexible mechanism for test-time performance improvement, enabling a controllable trade-off between inference-speed and accuracy within a single trained model.

Compared to the diffusion-based approach DiffusionDet \citep{chen2023diffusiondet} \squared{15}, which models diffusion only over bounding boxes, our approach \squared{17} provides a $2.6$ AP improvement under identical number of number of box refinement steps ($T^{\text{box}}=4$). This margin increases to $3.9$ AP when additionally running the label diffusion \squared{19}, thus highlighting the effectiveness of our proposed factorization that models labels and boxes separately.

It should be noted that although recent state-of-the-art methods like RT-DETR \citep{zhao2024detrs} \squared{21} - \squared{22} and DEIM-RT-DETR \citep{huang2025deim} \squared{23} report higher absolute numbers, these methods utilize specialized architectural components -- such as hybrid encoders and optimized matching strategies. As MDM provides a general diffusion-based framework for modeling structured tasks, these design choices are orthogonal to our formulation and can be incorporated into a specific instantiation of MDM to improve performance.

\subsection{Scene Graph Generation}
\label{sec:exp-sg}
\input{tables/vg-scenegraph}
\textbf{Dataset.} We evaluate performance on the Visual Genome benchmark dataset \citep{krishna2017visual}, using the widely adopted preprocessed subset introduced by \citet{xu2017scene}. This subset comprises $108$k images, spanning $150$ object categories and $50$ relation labels. 

\textbf{Evaluation Protocol.} To measure performance, we report results on standard scene graph evaluation metrics, namely Recall (R@K) \citep{xu2017scene}, Mean Recall (mR@K) \citep{chen2019knowledge,tang2019learning}, and Harmonic Recall (hR@K) \citep{khandelwal2022iterative}. Additionally, we also report the mR@100 performance on each long-tail category subset, namely \emph{head}, \emph{body}, and \emph{tail} \citep{li2021bipartite}. R@K is computed in a class agnostic manner, whereas mR@K measures the average recall across predicate classes independently. However, as noted in \cite{khandelwal2022iterative}, there exists an inherent tradeoff between R@K and mR@K -- mR@K can be improved at the expense of R@K, and vice versa, depending on the strength of the biasing towards long-tail predicate classes. Existing approaches typically control this trade-off through data sampling \citep{desai2021learning, li2021bipartite} or loss reweighting \citep{yan2020pcpl, khandelwal2022iterative} techniques. However, these biasing mechanisms are fixed during training, making the trade-off difficult to adjust without additional optimization. 

To avoid expensive retraining, we instead adopt a simple test-time biasing strategy \citep{menon2020long} that enables flexible control over the trade-off between R@K and mR@K at inference time. Specifically, let $\widetilde{\mathbf{l}}^p \in \mathbb{R}^C$ be the unnormalized logits over predicate classes. Following \cite{khandelwal2022iterative}, we define a class weight $w_c$ as $\max\left\{(\nicefrac{\eta}{f_c})^\delta, 1.0 \right\}$ where $f_c$ is the frequency of the predicate class $c \in C$ in the training set, and $\eta, \delta$ are scaling parameters. Following the best performing variants in \cite{khandelwal2022iterative, kim2024groupwise, fu2025hybrid}, we set $\eta=0.07,\delta=0.75$. Intuitively, less frequent predicate classes are assigned larger weights, thereby compensating for the long-tail imbalance in the data distribution. We then define the biased distribution $\mathbf{l}^p$ as,
\begin{align}
    \label{eq:sg-scaling}
    \mathbf{l}^p[c] = \sigma\left( \widetilde{\mathbf{l}}^p[c] + \lambda\log w_c \right)
\end{align}
where $\lambda$ controls the biasing strength. Increasing $\lambda$ amplifies the contribution of infrequent predicate classes, thus improving mR@K at the cost of R@K, while smaller values favor frequent predicates, leading to higher overall recall. Importantly, since this adjustment is applied directly to the predicted logits at inference time, the trade-off can be tuned without any additional training or modification to the learned model parameters.

\textbf{Implementation Details.} Scene graph generation can be formulated as predicting a set of \texttt{<subject, object, predicate>} triplets. Accordingly, recent methods \citep{khandelwal2022iterative, kim2021hotr, fu2025hybrid} adopt transformer-based architectures in which each component of the triplet is predicted by a separate decoder. To ensure a fair comparison with prior works we set $Q_{\text{train}} = 300$, use a ResNet-101 backbone \citep{he2016deep} as the feature extractor in the image encoder, and instantiate the MDM framework with a denoising decoder consisting of three separate decoders for subject, object, and predicate, each with $6$ layers. It should be noted that architecturally each decoder is similar to the formulation described in Section \ref{sec:tranformer-denoising}, and we do not introduce any task-specific components. We provide additional details in the Appendix (Section \ref{appendix:sg-arch}).

\textbf{Results.} As highlighted in Table \ref{tab:visualgenome}, MDM \tightsquare{27} outperforms deterministic baselines on the Visual Genome test set, demonstrating strong performance across head, body, and tail categories. Compared to baselines \tightsquare{1}–\tightsquare{12} using the stronger ResNeXt-101-FPN backbone \citep{xie2017aggregated}, MDM \tightsquare{27} outperforms the best baseline in \cite{li2024leveraging} \tightsquare{12} by $4.1/4.1$ on hR@50/100. When compared to baselines \tightsquare{16}–\tightsquare{22} using the same ResNet-101 backbone \citep{he2016deep}, MDM \tightsquare{27} consistently achieves better overall performance, improving over the strong baseline in \cite{kim2024groupwise} \tightsquare{22} by $3.3/4.1$ on hR@50/100. 

It should be noted that the most competitive baseline in HQSG \cite{fu2025hybrid} \tightsquare{23} is trained with $600$ input queries ($Q_{\text{train}} = 600$). As highlighted in Section \ref{sec:mscoco-results}, MDM can seamlessly function across a wide range of input queries without any additional training. As such, when evaluated under the $600$ query setting, MDM \tightsquare{29} outperforms the baseline in HQSG, achieving a $2.6/1.6$ improvement on hR@50/100.

Additionally, existing approaches such as ISG \tightsquare{21}, SpeaQ$_\text{ISG}$ \tightsquare{22}, and HQSG \tightsquare{23} utilize training-time strategies to facilitate the trade off between recall and mean recall. As such, making any changes to this trade off requires expensive retraining. MDM, on the other hand, can employ the simple test-time biasing strategy described in \ref{eq:sg-scaling} to allow effectively function across a wide range of recall and mean recall values without additional optimization (\tightsquare{31} - \tightsquare{35}). Specifically, by tuning the parameter $\lambda$ during inference, a single trained MDM model can biased to maximize performance on the rare body and tail categories \tightsquare{35} ($\lambda = 0.5$) or shifted to favor frequent head classes \tightsquare{32} ($\lambda = 0.1$).

\subsection{Qualitative Analysis}
\label{sec:qualitative-analysis}
MDM models the joint distribution over structured outputs, allowing it to produce multiple plausible predictions given a single input image.We illustrate this property for instance segmentation in Figure \ref{fig:inst-seg} by comparing outputs from our proposed MDM with those from the deterministic Mask R-CNN model \citep{he2017mask}. For MDM, given an input image, we draw $1000$ samples and aggregate the resulting masks into a heatmap that reflects prediction variability. Regions of higher intensity indicate areas that are consistently predicted across samples, while darker regions highlight uncertainty. We also visualize variation in the predicted bounding boxes by plotting the mean box for each object and shading the region spanned by the boxes with the minimum and maximum areas across samples. The extent of this shaded region reflects the variability in the predictions, with larger regions indicating greater uncertainty in the object’s localization. 

Contrary to deterministic approaches that generate a fixed prediction, MDM is able to identify challenging instances in at least a subset of samples. Subsequently, it is able to provide an estimate of uncertainty by aggregating information across samples. For example, in Figure \ref{fig:inst-seg-a}, the ``bird'' on the right is partially occluded by tree leaves. MDM reflects this ambiguity through uncertainty in both localization (the \textcolor{green}{green} shaded region), and shape (darker regions in the segmentation heatmap). Similar behavior is observed in Figures \ref{fig:inst-seg-b} (the ``giraffe'') and \ref{fig:inst-seg-c} (the ``skis'').

This diversity is more pronounced for the task of scene graph generation, where the combinatorial output space allows for many valid interpretations. As a result, MDM is able to generate distinct but plausible graphs given the same input image. We highlight some examples in Figure \ref{fig:sg}.

\subsection{Ablation Study}
\input{tables/ablation}
\begin{figure}[t]
    \centering
    \includegraphics[width=0.8\linewidth]{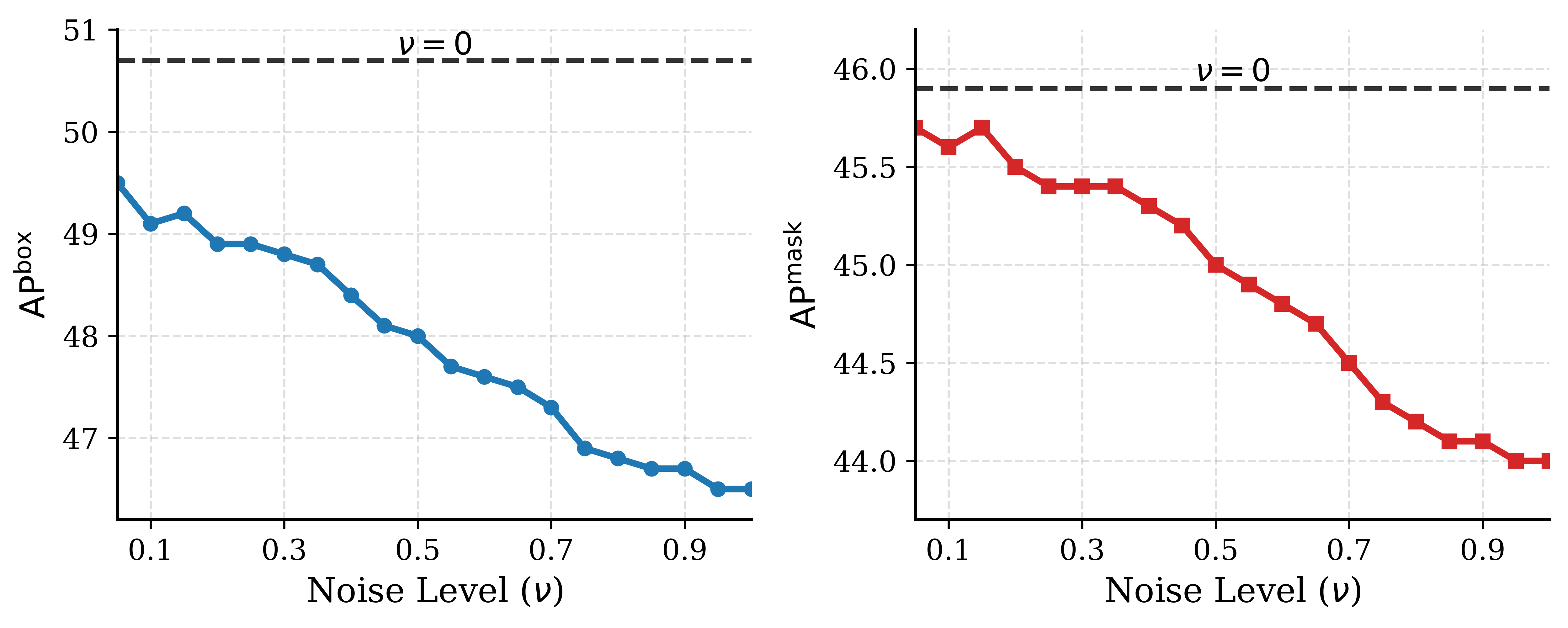}
    \caption{
    \textbf{Robustness to Noise in Conditioning}. \textbf{AP$^\text{box}$} (left) and  \textbf{AP$^\text{mask}$} (right) results shown for the MDM model trained on the instance segmentation task. We progressively corrupt conditioning signals using the forward diffusion process with noise level $\nu \in [0,1]$. Here, $\nu = 0$ corresponds to clean conditioning, while $\nu = 1$ corresponds to fully noised inputs drawn from $\mathcal{N}(\mathbf{0}, \mathbf{I})$. Performance degrades smoothly under increasing corruption, indicating that downstream diffusion modules remain robust to errors propagated from earlier stages of ancestral inference.
    }
    \label{fig:noise-robustness}
\end{figure}

\textbf{Drawbacks of Joint Diffusion Modeling.} For structured prediction tasks, MDM factorizes the joint conditional distribution into a sequence of simpler conditionals, each modeled by a separate diffusion process. As discussed in Section \ref{sec:intro}, an alternative is to model the full joint distribution using a single diffusion process, where all components are corrupted simultaneously, fed into a denoising network, and jointly denoised.

We empirically compare this joint formulation with MDM in Table \ref{tab:coco-inst-ablation} on the instance segmentation task. Both approaches use the transformer-based architecture described in Section~\ref{sec:tranformer-denoising}, consisting of a 6-layer encoder and a 9-layer denoising decoder. Importantly, as noted in Section \ref{sec:training}, MDM has a comparable number of parameters to the joint diffusion model due to parameter sharing.

As shown in Table \ref{tab:coco-inst-ablation}, the joint diffusion formulation consistently underperforms MDM across all metrics, and its performance further degrades as the number of diffusion steps increases (\eg box AP drops from 46.9 to 45.7, and mask AP from 43.1 to 42.2 when increasing $T$ from 1 to 4). In contrast, MDM (i) achieves a substantially higher performance (50.7 box AP and 45.9 mask AP), and (ii) shows continual improvement with more diffusion steps. This corroborates our hypothesis that jointly corrupting all components breaks the alignment between boxes, labels, and masks, leading to incoherent intermediate states that are difficult to denoise. As a result, the joint formulation is unable to accurately capture the underlying structured data distribution, as is evidenced by its poor performance with additional denoising steps. MDM, on the other hand, preserves cross-component consistency through its conditional factorization, allowing for improved modeling of the structured output space.

\textbf{Controllable Inference with MDMs.} As MDMs learn a separate diffusion process for each component of a structured representation, during inference, the number of times each diffusion process is run can be independently controlled. As such, this allows flexibility and fine-grained controlled over the generation process without any additional training. We highlight this in Tables \ref{tab:coco-inst-results}, \ref{tab:coco-det-results}, and \ref{tab:visualgenome}. 

Although MDM is capable of producing valid structured outputs from each individual diffusion process, chaining these processes together leads to improved performance. For example, in instance segmentation (Table \ref{tab:coco-inst-results}), while the box diffusion process by itself is competitive, we observe steady gains when progressively incorporating additional components. Adding label diffusion improves performance by $0.8$ AP$^\text{mask}$ and $1.4$ AP$^\text{box}$ (\circled{9} $\rightarrow$ \circled{11}), and further including mask diffusion yields additional gains of $0.1$ on AP$^\text{mask}$ and $0.2$ AP$^\text{box}$ (\circled{11} $\rightarrow$ \circled{13}). This further highlights the complementary nature and benefits of modeling each component of the structured space as an individual diffusion process.

\textbf{Impact of Number of Queries.} As discussed in Section \ref{sec:training}, our model is trained with a fixed number of queries ($Q_{\text{train}}$). Unlike deterministic approaches, which require the number of inference queries to match the training setup ($Q_{\text{eval}} = Q_{\text{train}}$), MDMs enable flexible adjustment of the query count without requiring any additional training. As such, we vary the number of queries ($Q_{\text{eval}}$) for each task and report the results in Tables \ref{tab:coco-inst-results}, \ref{tab:coco-det-results}, and \ref{tab:visualgenome}.

The impact of increasing $Q_{\text{eval}}$ depends strongly on the structure and complexity of the output space. For problems with relatively constrained output spaces -- such as object detection and instance segmentation -- where the number of valid predictions scales linearly with the number of objects, a limited number of queries is sufficient to cover most plausible outputs.  For example, increasing $Q_{\text{eval}}$ from 100 to 300 yields moderate improvements, with gains of 1.1 AP for object detection (\squared{20} $\rightarrow$ \squared{19}) and 0.8 AP$^{\text{mask}}$ for instance segmentation (\circled{14} $\rightarrow$ \circled{13}). Beyond this performance quickly saturates, and further increasing the number of queries provides diminishing returns, with negligible improvements observed beyond $Q_{\text{eval}} = 600$.

In contrast, scene graph generation is a fundamentally more complex problem, where the output space is combinatorial in the number of objects. In this setting, increasing $Q_{\text{eval}}$ substantially improves coverage of plausible relations triplets. For example, going from 100 to 300 queries results in gains of 0.9/1.6 in mR@50/100 and 2.0/2.6 in R@50/100 (\tightsquare{28} $\rightarrow$ \tightsquare{27}). This trend continues as $Q_{\text{eval}}$ is increased further, where going from 300 to 900 queries provides additional improvements of 0.5/0.6 in mR@50/100 and 1.3/1.5 in R@50/100 (\tightsquare{27} $\rightarrow$ \tightsquare{30}).

Importantly, while increasing $Q_{\text{eval}}$ improves coverage in complex output spaces, it also increases inference runtime. However, this overhead is sublinear, as a substantial portion of computation is independent of the number of queries (image encoder). In particular, increasing $Q_{\text{eval}}$ from $100$ to $900$ makes inference approximately $1.3\times$ slower.

\textbf{Robustness to Noise in Ancestral Conditioning.} As described in Section \ref{sec:inference}, inference in MDM follows an ancestral sampling procedure, where each diffusion module conditions on the outputs of its parent processes. As a result, errors introduced in earlier stages of the sampling process may propagate to subsequent modules. To explicitly study this effect, we perform a controlled stress test in which the conditioning inputs to each diffusion process are deliberately corrupted. Concretely, we apply the forward diffusion process defined in Eq. (\ref{eq:gaussianforwardprocessx0}) to the conditioning variables and perturb them to an intermediate noise level corresponding to timestep $\nu T$, where $T$ is the total number of diffusion steps and $\nu \in [0,1]$ controls the severity of corruption. Note that $\nu = 0$ corresponds to no corruption of the conditioning, while $\nu = 1$ corresponds to fully noised conditioning, \ie samples drawn from $\mathcal{N}(\mathbf{0}, \mathbf{I})$.

We highlight the results for instance segmentation in Figure \ref{fig:noise-robustness}. Across both box and mask AP, MDM exhibits a smooth and gradual degradation in performance as $\nu$ increases, rather than an abrupt collapse. Each diffusion stage is therefore able to recover meaningful structure even when conditioned on heavily corrupted inputs. Notably, even at $\nu = 1$, where the conditioning is fully noised and drawn from $\mathcal{N}(\mathbf{0}, \mathbf{I})$, MDM retains a substantial fraction of its performance, indicating strong robustness to severe upstream corruption in the ancestral inference process.

%% file: tables/coco-instance-segmentation.tex
\begin{table}[t]
\resizebox{\linewidth}{!}{
\centering
\begin{tabular}{l|c|l|>{\centering\arraybackslash}m{1.2cm}|>{\centering\arraybackslash}m{1.2cm}|>{\centering\arraybackslash}m{1.2cm}|>{\centering\arraybackslash}m{1.2cm}|cc|ccccc}
\toprule
\toprule
& \textbf{\#} & \multicolumn{5}{l|}{\textbf{Method}} & \textbf{AP$^\text{mask}$} & \textbf{AP$^\text{box}$}& \textbf{AP$^\text{mask}_{50}$} & \textbf{AP$^\text{mask}_{75}$} & \textbf{AP$^\text{mask}_s$} & \textbf{AP$^\text{mask}_m$} & \textbf{AP$^\text{mask}_l$} \\ 
\cmidrule{1-14} 
\multirow{5}{*}{\rotatebox[origin=c]{90}{Deterministic}} 
& \circled{1} & \multicolumn{5}{l|}{Mask-RCNN \citep{he2017mask}} & 42.5 & 48.2 & - & - & 23.8 & 45.0 & 60.0 \\
& \circled{2} & \multicolumn{5}{l|}{HTC \citep{chen2019hybrid}} & 39.7 & 44.9 & 61.4 & 43.1 &  22.6 & 42.2 & 50.6 \\
& \circled{3} & \multicolumn{5}{l|}{QueryInst \citep{fang2021instances}} & 40.6 & 45.6 & 63.0 & 44.0 & 23.4 & 42.5 & 52.8 \\
& \circled{4} & \multicolumn{5}{l|}{Mask2Former \citep{cheng2022masked}} & 43.7 & 46.2 & 66.0 & 46.9 & 23.4 & 47.2 & 64.8 \\
& \circled{5} & \multicolumn{5}{l|}{Mask DINO \citep{li2023mask}} & \textbf{46.0} & 50.5 & 68.9 & \textbf{50.3} & 26.0 & \textbf{49.3} & 65.5 \\

\cmidrule{1-14} 

\multirow{14}{*}{\rotatebox[origin=c]{90}{Diffusion}}
& \circled{6} & \multicolumn{5}{l|}{DiffusionInst; $T=1$ \citep{gu2024diffusioninst}} & 37.3 & - & 60.3 & 39.3 & 18.9 & 40.1 & 54.7 \\
& \circled{7} & \multicolumn{5}{l|}{DiffusionInst; $T=4$ \citep{gu2024diffusioninst}} & 37.5 & - & 60.9 & 39.3 & 19.2 & 40.4 & 54.8 \\
\cmidrule{2-14} 

&  & \multirow{11}{*}{MDM} & $Q_{\text{eval}}$  & $T^{\text{box}}$ & $T^{\text{label}}$ & $T^{\text{mask}}$ &  &  &  &  &  &  & \\
\cmidrule{2-7}
& \circled{8} &  & \multirow{6}{*}{$300$}  & 1 & - & - & 45.0 & 49.1 & 67.7 & 48.9 & 25.0 & 48.4 & 65.8 \\
& \circled{9} &  &  & 4 & - & - & 44.9 & 49.1 & 68.1 & 48.5 & 25.0 & 48.3 & 65.1 \\
& \circled{10} &  &  & 4 & 1 & - & 45.7 & 50.3 & 68.9 & 49.7 & 26.1 & 49.0 & 65.8 \\
& \circled{11} &  &  & 4 & 4 & - & 45.8 & 50.5 & 68.9 & 49.8 & 26.2 & 49.1 & 66.0 \\
& \circled{12} &  &  & 4 & 4 & 1 & 45.8 & 50.6 & \underline{68.9} & 49.8 & \underline{26.4} & 49.1 & \underline{66.0} \\
& \circled{13} &  &  & 4 & 4 & 4 & \underline{45.9} & 50.7 & \textbf{68.9} & 49.9 & \textbf{26.5} & 49.1 & \textbf{66.1} \\
\cmidrule{4-14} 
& \circled{14} &  & $100$ & \multirow{3}{*}{4} & \multirow{3}{*}{4} & \multirow{3}{*}{4} & 45.1 & 49.5 &  68.0 & 48.8 & 25.5 & 48.4 & 65.8 \\
& \circled{15} &  & $600$ &  &  &  & 45.9 & \textbf{50.9} & 68.8 & \underline{50.1} & 26.1 & \underline{49.2} & 65.9 \\
& \circled{16} &  & $900$ &  &  &  & 45.8 & \underline{50.8} & 68.6 & 49.9 & 26.0 & 49.1 & 65.7 \\

\bottomrule 
\bottomrule 
\end{tabular}%
}
\caption{\textbf{Instance segmentation.} Results on MS-COCO \emph{val2017} with ResNet-50 backbone. The best results are shown in \textbf{bold}, and the second-best results are \underline{underlined}. MDM \protect\circled{13} achieves performance on par with the strong baseline \protect\circled{5}. Note that although MDM is trained with $300$ queries, it can operate over different query budgets ($Q_{\text{eval}})$ \protect\circled{14}-\protect\circled{16} without any additional training.}
\label{tab:coco-inst-results}
\end{table}

%% file: tables/coco-obj-detection.tex
\begin{table}[t]
\resizebox{\linewidth}{!}{
\centering
\begin{tabular}{l|c|>{\raggedright\arraybackslash}m{4.0cm}|>{\centering\arraybackslash}m{1.0cm}|>{\centering\arraybackslash}m{1.0cm}|>{\centering\arraybackslash}m{1.0cm}|cccccc}
\toprule
\toprule
& \textbf{\#} & \multicolumn{4}{l|}{\textbf{Method}} & \textbf{AP} & \textbf{AP$_{50}$} & \textbf{AP$_{75}$} & \textbf{AP$_s$} & \textbf{AP$_m$} & \textbf{AP$_l$} \\ 
\cmidrule{1-12} 
\multirow{16}{*}{\rotatebox[origin=c]{90}{Deterministic}} 
& \squared{1} & \multicolumn{4}{l|}{RetinaNet \citep{lin2017focal}} & 38.7 & 58.0 & 41.5 & 23.3 & 42.3 & 50.3 \\
& \squared{2} & \multicolumn{4}{l|}{Faster R-CNN \citep{ren2015faster}} & 40.2 & 61.0 & 43.8 & 24.2 & 43.5 & 52.0 \\
& \squared{3} & \multicolumn{4}{l|}{DETR \citep{carion2020end}} & 42.0 & 62.4 & 44.2 & 20.5 & 45.8 & 61.1 \\
& \squared{4} & \multicolumn{4}{l|}{Conditional-DETR \citep{meng2021conditional}} & 43.0 & 64.0 & 45.7 & 22.7 & 46.7 & 61.5 \\
& \squared{5} & \multicolumn{4}{l|}{Cascade R-CNN \citep{wang2022anchor}} & 44.2 & 64.7 & 47.5 & 24.7 & 48.2 & 60.6\\
& \squared{6} & \multicolumn{4}{l|}{Cascade R-CNN \citep{cai2018cascade}} & 44.3 & 62.2 & 48.0 & 26.6 & 47.7 & 57.7 \\
& \squared{7} & \multicolumn{4}{l|}{Sparse R-CNN \citep{sun2021sparse}} & 45.0 & 63.4 & 48.2 & 26.9 & 47.2 & 59.5 \\
& \squared{8} & \multicolumn{4}{l|}{Efficient-DETR \citep{yao2021efficient}} & 45.1 & 63.1 & 49.1 & 28.3 & 48.4 & 59.0 \\
& \squared{9} & \multicolumn{4}{l|}{SMCA-DETR \citep{gao2021fast}} & 45.6 & 65.5 & 49.1 & 25.9 & 49.3 & 62.6 \\
& \squared{10} & \multicolumn{4}{l|}{Deformable-DETR \citep{zhu2020deformable}} & 46.2 & 65.2 & 50.0 & 28.8 & 49.2 & 61.7\\
& \squared{11} & \multicolumn{4}{l|}{DAB-DETR$^\dagger$ \citep{liu2022dab}} & 46.9 & 66.0 & 50.8 & 30.1 & 50.4 & 62.5 \\
& \squared{12} & \multicolumn{4}{l|}{DN-DETR$^\dagger$ \citep{li2022dn}} & 48.6 & 67.4 & 52.7 & 31.0 & 52.0 & 63.7 \\
& \squared{13} & \multicolumn{4}{l|}{DINO-DETR$^\dagger$ \citep{zhang2022dino}} & \textbf{50.9} & 69.0 & 55.3 & \underline{34.6} & \textbf{54.1} & 64.6 \\

\cmidrule{1-12} 

\multirow{12}{*}{\rotatebox[origin=c]{90}{Diffusion}}
& \squared{14} & \multicolumn{4}{l|}{DiffusionDet; $T=1$ \citep{chen2023diffusiondet}} & 45.8 & 64.1 & 50.4 & 27.6 & 48.7 & 62.2 \\
& \squared{15} & \multicolumn{4}{l|}{DiffusionDet; $T=4$ \citep{chen2023diffusiondet}} & 46.6 & 65.1 & 51.3 & 28.9 & 49.2 & 62.1 \\

\cmidrule{2-12} 

&  & \multirow{11}{*}{MDM} & $Q_{\text{eval}}$ & $T^{\text{box}}$ & $T^{\text{label}}$  &  &  &  &  \\
\cmidrule{2-6}
& \squared{16} &  & \multirow{4}{*}{$300$} &  1 & - & 49.3 & 67.7 & 53.9 & 32.9 & 52.8 & 64.0 \\ 
& \squared{17} &  &  &  4 & - & 49.2 & 68.9 & 52.9 & 33.2 & 52.1 & 63.5 \\ 
& \squared{18} &  &  &  4 & 1 & 50.3 & \textbf{69.4} & 54.7 & 34.5 & 53.5 & 64.4 \\
& \squared{19} &  &  &  4 & 4 & 50.5 & \underline{69.2} & 55.0 & 34.3 & 53.5 & 64.8 \\ 
\cmidrule{4-12} 
& \squared{20} &  & $100$ & \multirow{3}{*}{4} & \multirow{3}{*}{4}  & 49.4 & 68.1 & 53.6 & 33.5 & 52.2 & 63.9 \\
& \squared{21} &  & $600$ &  &   & 50.7 & 69.2 & \underline{55.4} & 34.5 & 53.8 & \textbf{65.1} \\
& \squared{22} &  & $900$ &  &  & \underline{50.7} & 69.0 & \textbf{55.5} & \textbf{34.7} & \underline{53.9} & \underline{64.9} \\

\cmidrule{1-12}
\multicolumn{11}{c}{Not Directly Comparable} \\
\cmidrule{1-12}
\rowcolor{gray!30}
& \textcolor{black!50}{\squared{23}} & \multicolumn{4}{l|}{\textcolor{black!50}{RT-DETR \citep{zhao2024detrs}}} & \textcolor{black!50}{53.1} & \textcolor{black!50}{71.3} & \textcolor{black!50}{57.7} & \textcolor{black!50}{34.8} & \textcolor{black!50}{58.0} & \textcolor{black!50}{70.0} \\
\rowcolor{gray!30}
& \textcolor{black!50}{\squared{24}} & \multicolumn{4}{l|}{\textcolor{black!50}{RT-DETRv2 \citep{lv2024rt}}} & \textcolor{black!50}{53.4} & \textcolor{black!50}{71.6} & \textcolor{black!50}{57.4} & \textcolor{black!50}{36.1} & \textcolor{black!50}{57.9} & \textcolor{black!50}{70.8} \\
\rowcolor{gray!30}
& \textcolor{black!50}{\squared{25}} & \multicolumn{4}{l|}{\textcolor{black!50}{DEIM-RT-DETR \citep{huang2025deim}}} & \textcolor{black!50}{54.3} & \textcolor{black!50}{72.3} & \textcolor{black!50}{58.8} & \textcolor{black!50}{37.5} & \textcolor{black!50}{58.7} & \textcolor{black!50}{70.8} \\
\bottomrule 
\bottomrule 
\end{tabular}%
}
\caption{\textbf{Object detection.} Results on MS-COCO \emph{val2017} with a ResNet-50 backbone. $^\dagger$ denotes models trained with an effective query count of $900$. Unlike deterministic methods that require retraining to increase the number of queries, MDM trained with $300$ queries can be evaluated with $900$ queries \protect\squared{22} without retraining. The best results are shown in \textbf{bold}, and the second-best results are \underline{underlined}. Methods in \protect\squared{23}–\protect\squared{25} are not directly comparable, as they rely on specialized architectural components that are orthogonal to our formulation. Notably, these components can be incorporated into MDM to provide further improvements.}
\label{tab:coco-det-results}
\end{table}

%% file: tables/vg-scenegraph.tex
\begin{table}[t]
\centering
\setlength{\tabcolsep}{3.0pt}
\resizebox{\linewidth}{!}{%
\begin{tabular}{l|l|c|>{\raggedright\arraybackslash}m{5.0cm}|>{\centering\arraybackslash}m{1.0cm}|>{\centering\arraybackslash}m{1.0cm}|>{\centering\arraybackslash}m{1.0cm}|>{\centering\arraybackslash}m{1.0cm}|ccc|ccc}
\toprule
\toprule
                                  & \textbf{B}             &\textbf{\#} & \multicolumn{5}{l|}{\textbf{Method}}     & \textbf{mR@50/100}       & \textbf{R@50/100}  & \textbf{hR@50/100}        & \textbf{Head} & \textbf{Body} & \textbf{Tail} \\ \cmidrule{1-14} 
                                  \multirow{23}{*}{\rotatebox[origin=c]{90}{Deterministic}}   & \multirow{12}{*}{\rotatebox[origin=c]{90}{X101-FPN}} & \tightsquare{1} & \multicolumn{5}{l|}{RelDN \citep{zhang2019graphical}}             & ${6.0} ~/~ {7.3}$    & ${31.4}~/~{35.9}$ & $10.1~/~12.1$ & -             & -             & -             \\
                                                              &                               &  \tightsquare{2} & \multicolumn{5}{l|}{MOTIF \citep{zellers2018neural}}               & ${5.5}~/~{6.8}$   & ${32.1}~/~{36.9}$ & $9.4 ~/~ 11.5$ & -             & -             & -             \\
                                                              &                               & \tightsquare{3} & \multicolumn{5}{l|}{VCTree \citep{tang2019learning}}             & ${6.6}~/~{7.7}$   & ${31.8}~/~{36.1}$ & $10.9 ~/~ 12.7$ &  -             & -             & -             \\
                                                              &                               & \tightsquare{4} & \multicolumn{5}{l|}{BGNN \citep{li2021bipartite}}              & ${10.7}~/~{12.6}$ & ${31.0}~/~{35.8}$ & $15.9 ~/~ 18.6$ & $\textbf{34.0}$        & $12.9$        & $6.0$         \\ 
                                                              &                               & \tightsquare{5} & \multicolumn{5}{l|}{VCTree-TDE \citep{tang2020unbiased}}          & ${9.3}~/~{11.1}$  & ${19.4}~/~{23.2}$  & $12.6 ~/~ 15.0$ & -             & -             & -             \\
                                                              &                               & \tightsquare{6} & \multicolumn{5}{l|}{VCTree-DLFE \citep{chiou2021recovering}}         & ${11.8}~/~{13.8}$  & ${22.7}~/~{26.3}$  & $15.5 ~/~ 18.1$ & -             & -             & -             \\
                                                              &                               & \tightsquare{7} & \multicolumn{5}{l|}{VCTree-EBM \citep{suhail2021energy}}         & ${9.7}~/~{11.6}$   & ${20.5}~/~{24.7}$  & $13.2 ~/~ 15.8$ & -             & -             & -             \\
                                                              &                               & \tightsquare{8} & \multicolumn{5}{l|}{VCTree-BPLSA \citep{guo2021general}}       & ${13.5}~/~{15.7}$  & ${21.7}~/~{25.5}$  & $16.6 ~/~ 19.4$ & -             & -             & -             \\
                                                              &                               & \tightsquare{9} & \multicolumn{5}{l|}{DT2-ACBS \citep{desai2021learning}}           & $\textbf{22.0}~/~\textbf{24.4}$  & ${15.0}~/~{16.3}$  & $17.8 ~/~ 19.5$ &  -             & -             & -             \\ 
                                                              &                               & \tightsquare{10} & \multicolumn{5}{l|}{PE-Net \citep{zheng2023prototype}} &  ${12.4}~/~{14.5}$ & ${30.7}~/~{35.2}$  &  ${17.7}~/~{20.5}$ & - & - & - \\ 
                                                              &                               & \tightsquare{11} & \multicolumn{5}{l|}{VETO \citep{sudhakaran2023vision}} & ${10.6}~/~{13.8}$ & ${28.6}~/~{34.0}$  & ${15.5}~/~{19.6}$ & - & - & - \\    
                                                              &                               & \tightsquare{12} & \multicolumn{5}{l|}{DRM \citep{li2024leveraging}} & ${20.4}~/~\underline{24.1}$  & ${19.0}~/~{22.9}$ & ${19.7}~/~{23.5}$ & - & - & - \\
                                                              \cmidrule{2-14} 
                                   &   \multirow{3}{*}{\rotatebox[origin=c]{90}{{\shortstack{ResNet\\50}}}}          & \tightsquare{13} & \multicolumn{5}{l|}{RelTR \citep{cong2023reltr}} & ${10.8}~/~{12.3}$ & ${27.5}~/~{30.7}$ & ${15.5}~/~{17.6}$ & - & - & - \\
                                                              &                               & \tightsquare{14} & \multicolumn{5}{l|}{EGTR \citep{im2024egtr}} & ${17.8}~/~{21.7}$ & ${18.7}~/~{20.5}$ & ${18.2}~/~{21.1}$ & - & - & - \\
                                                              &                               & \tightsquare{15} & \multicolumn{5}{l|}{Hydra-SGG \citep{chen2024hydra}} & ${15.9}~/~{19.4}$ & ${28.6}~/~{33.4}$ & ${20.5}~/~{24.7}$ & - & - & - \\
                                                              \cmidrule{2-14} 
                                   &    \multirow{8}{*}{\rotatebox[origin=c]{90}{ResNet-101}}                           & \tightsquare{16} & \multicolumn{5}{l|}{BGNN \citep{li2021bipartite,li2021sgtr}}               & ${8.6}~/~{10.3}$   & ${28.2}~/~{33.8}$  & $13.2 ~/~ 15.8$ &  $29.1$        & $12.6$        & $2.2$         \\
                                                              &                               & \tightsquare{17} & \multicolumn{5}{l|}{RelDN \citep{zhang2019graphical,li2021sgtr}}               & ${4.4}~/~{5.4}$    & {${30.3}~/~{34.8}$}  & $7.7 ~/~ 9.3$ & $31.3$        & $2.3$         & $0.0$         \\
                                                              &         & \tightsquare{18} & \multicolumn{5}{l|}{AS-Net \citep{chen2021reformulating}}             & ${6.1}~/~{7.2}$   & ${18.7}~/~{21.1}$  & $9.2 ~/~ 10.7$ & $19.6$        & $7.7$         & $2.7$         \\
                                                              &                               & \tightsquare{19} & \multicolumn{5}{l|}{HOTR \citep{kim2021hotr}}                & ${9.4}~/~{12.0}$   & ${23.5}~/~{27.7}$  & $13.4 ~/~ 16.7$ & $26.1$        & $16.2$        & $3.4$         \\
                                                              &                               & \tightsquare{20} & \multicolumn{5}{l|}{SGTR \citep{li2021sgtr}} & ${15.8}~/~{20.1}$  & ${20.6}~/~{25.0}$  & $17.9 ~/~ 22.3$ & $21.7$        & {$21.6$}        & 
                                                              ${17.1}$        \\
                                                              &   & \tightsquare{21} &  \multicolumn{5}{l|}{ISG$^*$ \citep{khandelwal2022iterative}} & $15.7 ~/~ 17.8$ & $27.2 ~/~ 29.8$ & {$19.9 ~/~ 22.3$} &$28.5$ & $18.8$ & ${13.3}$     \\   

                                                              &   & \tightsquare{22} &  \multicolumn{5}{l|}{SpeaQ$_\text{ISG}^{*}$ \citep{kim2024groupwise}} & $15.1 ~/~ 17.6$ & $32.1 ~/~ 35.5$ & {$20.5 ~/~ 23.5$} &- & - & -     \\   
                                                              &   & \tightsquare{23} &  \multicolumn{5}{l|}{HQSG$^{* \ddagger}$ \citep{fu2025hybrid}} & $16.0 ~/~ 20.5$ & $\textbf{34.1} ~/~ \textbf{38.3}$ & {$21.8 ~/~ 26.7$} &- & - & -     \\   
                                                            \cmidrule{1-14} 
                                \multirow{15}{*}{\rotatebox[origin=c]{90}{Diffusion}} & \multirow{15}{*}{\rotatebox[origin=c]{90}{ResNet-101}} &  & \multirow{15}{*}{MDM} & $Q_{\text{eval}}$  & $\lambda$ & $T^{\text{box}}$ & $T^{\text{label}}$  &  &  &  &  &  &  \\
                                \cmidrule{3-8}
                                 &  & \tightsquare{24} & & \multirow{4}{*}{$300$} & \multirow{4}{*}{0.4} &  1 & -  & $18.3 ~/~ 21.0$ & $29.2 ~/~ 32.8$  & $22.5 ~/~ 25.6$ & $28.4$ & $26.1$ & $13.8$ \\
                                 &  & \tightsquare{25} & & & & 4 & -  & $18.6 ~/~ 21.8$ & $30.2 ~/~ 34.3$ & $23.0 ~/~ 26.7$ & $29.3$ & $27.2$ & $14.3$ \\
                                 &  & \tightsquare{26} & & & & 4 & 1  & $19.3 ~/~ 22.7$ & $30.2 ~/~ 34.5$ & $23.5 ~/~ 27.4$ & $30.1$ & $28.2$ & $15.0$ \\
                                 &  & \tightsquare{27} & & & & 4 & 4  & $19.7 ~/~ 22.9$ & $30.2 ~/~ 34.6$  & $23.8 ~/~ 27.6$ & $30.2$ & $28.1$ & $15.6$ \\
                            \cmidrule{5 - 14}
                                &  & \tightsquare{28} &  & $100$ & & \multirow{3}{*}{4} & \multirow{3}{*}{4}  & $18.8 ~/~ 21.3$ & $28.2 ~/~ 32.0$ & $22.6 ~/~ 25.6$ & $27.9$ & $26.5$ & $14.3$ \\
                                &  & \tightsquare{29} &  & $600$ & & & & ${20.1} ~/~ {23.5}$ & $31.2 ~/~ {35.7}$ & $\underline{24.4} ~/~ \underline{28.3}$ & ${31.0}$ & $\underline{29.0}$ & ${15.8}$ \\
                                &  & \tightsquare{30} &  & $900$ & &  & & ${20.2} ~/~ {23.5}$ & ${31.5} ~/~ {36.1}$ & $\textbf{24.6} ~/~ \textbf{28.5}$ & ${31.2}$ & $\textbf{29.2}$ & $15.6$ \\
                            \cmidrule{5 - 14}
                                &  & \tightsquare{31} &  & \multirow{5}{*}{$300$} & 0.0 & \multirow{5}{*}{4} & \multirow{5}{*}{4}  & $13.5 ~/~ 15.9$ & $\underline{32.9} ~/~ \underline{37.3}$  & $19.1 ~/~ 22.3$ & $32.2$ & $23.6$ & $9.8$ \\
                                &  & \tightsquare{32} & & & 0.1 & &  & $15.3 ~/~ 17.9$ & $32.8 ~/~ 37.3$  & $20.9 ~/~ 24.2$ & $\underline{33.8}$ & $23.8$ & $7.2$ \\
                                &  & \tightsquare{33} & & & 0.2 & &  & $16.7 ~/~ 19.7$ & $32.3 ~/~ 36.7$  & $22.0 ~/~ 25.6$ & $32.9$ & $25.8$ & $9.7$ \\
                                &  & \tightsquare{34} & & & 0.3 & &  & $18.3 ~/~ 21.5$ & $31.6 ~/~ 36.1$  & $23.2 ~/~ 26.9$ & $31.8$ & $27.2$ & $12.6$ \\
                                &  & \tightsquare{35} & & & 0.5 & &  & $\underline{20.7} ~/~ 24.0$ & $27.7 ~/~ 31.7$  & $23.7 ~/~ 27.3$ & $27.2$ & $28.7$ & $\textbf{18.5}$ \\
                            
                            \bottomrule \bottomrule 
\end{tabular}%
}
\caption{\textbf{Scene Graph Generation on Visual Genome.} Mean Recall (mR@K), Recall (R@K), and Harmonic Recall (hR@K) shown for baselines and our approach on the Visual Genome \citep{krishna2017visual} test set. The best results are shown in \textbf{bold}, and the second-best results are \underline{underlined}. $^*$ denotes model trained using the loss-reweighting strategy proposed in \citep{khandelwal2022iterative}. $^\ddagger$ denotes model trained with $Q_{\text{train}} = 600$. MDM trained with $300$ queries can be evaluated with $600$ queries \protect\squared{29} without retraining.}
\label{tab:visualgenome}
\end{table}

%% file: tables/ablation.tex
\begin{table}[t]
\resizebox{\linewidth}{!}{
\centering
\begin{tabular}{l|>{\centering\arraybackslash}m{0.7cm}|>{\centering\arraybackslash}m{0.7cm}|>{\centering\arraybackslash}m{0.7cm}|cccccc|cccccc}
\toprule
\toprule
\multirow{2}{*}{Method} & \multicolumn{3}{c|}{\multirow{2}{*}{$T$}} & \multicolumn{6}{c|}{Box} & \multicolumn{6}{c}{Mask} \\
\cmidrule{5-16}
& \multicolumn{3}{c|}{} & \textbf{AP} & \textbf{AP$_{50}$} & \textbf{AP$_{75}$} & \textbf{AP$_s$} & \textbf{AP$_m$} & \textbf{AP$_l$} & \textbf{AP} & \textbf{AP$_{50}$} & \textbf{AP$_{75}$} & \textbf{AP$_s$} & \textbf{AP$_m$} & \textbf{AP$_l$} \\
\cmidrule{1-16}
\multirow{2}{*}{Joint Diffusion} & \multicolumn{3}{c|}{1} & 46.9 & 66.6 & 50.8 & 30.0 & 49.9 & 62.3 & 43.1 & 65.4 & 46.7 & 23.1 & 46.3 & 63.2 \\
 & \multicolumn{3}{c|}{4} & 45.7 & 65.9 & 49.1 & 29.4 & 48.4 & 61.6 & 42.2 & 64.6 & 45.4 & 22.7 & 45.7 & 62.5 \\
\cmidrule{1-16}
& $T^{\text{box}}$ & $T^{\text{label}}$ & $T^{\text{mask}}$ & & & & & & & & & & \\
\cmidrule{2-4}
\multirow{6}{*}{MDM} & 1 & - & - & 49.1 & 68.6 & 53.3 & 32.7 & 52.2 & 65.0 & 45.0 & 67.7 & 48.9 & 25.0 & 48.4 & 65.8 \\
& 4 & - & - & 49.1 & 69.5 & 52.7 & 32.5 & 51.9 & 64.3 & 44.9 & 68.1 & 48.5 & 25.0 & 48.3 & 65.1 \\
& 4 & 1 & - & 50.3 & 70.3 & 54.4 & 34.2 & 53.6 & 65.5 & 45.7 & 68.9 & 49.7 & 26.1 & 49.0 & 65.8 \\
& 4 & 4 & - & 50.5 & 70.2 & 54.8 & 34.1 & 53.6 & 66.3 & 45.8 & 68.9 & 49.8 & 26.2 & 49.1 & 66.0 \\
& 4 & 4 & 1 & \underline{50.6} & \textbf{70.3} & \underline{55.0} & \underline{35.3} & \underline{53.8} & \underline{66.3} & \underline{45.8} & \underline{68.9} & \underline{49.8} & \underline{26.4} & \underline{49.1} & \underline{66.0} \\
& 4 & 4 & 4 & \textbf{50.7} & \underline{70.2} & \textbf{55.0} & \textbf{35.4} & \textbf{53.8} & \textbf{66.5} & \textbf{45.9} & \textbf{68.9} & \textbf{49.9} & \textbf{26.5} & \textbf{49.1} & \textbf{66.1} \\
\bottomrule 
\bottomrule 
\end{tabular}%
}
\caption{\textbf{Drawbacks of Joint Modeling.} Results on MS-COCO \emph{val2017} with ResNet-50 backbone. Both formulations use a $6$-layer encoder, a $9$-layer denoising decoder, and $Q_{\text{train}} = Q_{\text{eval}} = 300$. In ``Joint Diffusion'', boxes, labels, and masks are corrupted simultaneously, and passed to the decoder to be denoised jointly. The best results are shown in \textbf{bold}, and the second-best results are \underline{underlined}. Note that MDM has a comparable number of parameters to the ``Joint Diffusion'' model. }
\label{tab:coco-inst-ablation}
\end{table}